\documentclass[journal]{IEEEtran}
\usepackage{url}
\usepackage{times}
\usepackage{epsfig}
\usepackage{graphicx}
\usepackage{amsmath}
\usepackage{amssymb}
\usepackage{caption}
\usepackage{subfigure}
\usepackage{booktabs}
\usepackage{caption}
\usepackage{stfloats}
\usepackage[dvipsnames]{xcolor}
\usepackage{soul}
\usepackage{siunitx}
\usepackage{hyperref}
\usepackage{xcolor}

\ifCLASSINFOpdf
\else
\fi
\begin{document}
%
\title{Improving the Harmony of the Composite Image \\ by Spatial-Separated Attention Module}
%
%
%

\author{Xiaodong~Cun
and Chi-Man~Pun,~\IEEEmembership{Senior~Member,~IEEE}\thanks{Xiaodong~Cun and Chi-Man~Pun are with the Department
of Computer and Information Science, University of Macau, Macau, 999078, China e-mail: \{yb87432,cmpun\}@umac.mo.}
\thanks{Corresponding author: Chi-Man~Pun}
\thanks{Code is available at: https://github.com/vinthony/s2am}
 }

%
%

\markboth{Journal of \LaTeX\ Class Files,~Vol.~14, No.~8, August~2015}%
{Shell \MakeLowercase{\textit{et al.}}: Bare Demo of IEEEtran.cls for IEEE Journals}
%



\maketitle

\begin{abstract}

Image composition is one of the most important applications in image processing. However, the inharmonious appearance between the spliced region and background degrade the quality of the image. Thus, we address the problem of Image Harmonization: Given a spliced image and the mask of the spliced region, we try to harmonize the ``style'' of the pasted region with the background (non-spliced region). 
Previous approaches have been focusing on learning directly by the neural network.
In this work, we start from an empirical observation: 
the differences can only be found in the spliced region between the spliced image and the harmonized result while they share the same semantic information and the appearance in the non-spliced region.
Thus, in order to learn the feature map in the masked region and the others individually, we propose a novel attention module named Spatial-Separated Attention Module (S$^{2}$AM). 
Furthermore, we design a novel image harmonization framework by inserting the S$^{2}$AM in the coarser low-level features of the Unet structure by two different ways. Besides image harmonization, we make a big step for harmonizing the composite image without the specific mask under previous observation.
The experiments show that the proposed S$^{2}$AM performs better than other state-of-the-art attention modules in our task. 
Moreover, we demonstrate the advantages of our model against other state-of-the-art image harmonization methods via criteria from multiple points of view.

\end{abstract}

%
\IEEEpeerreviewmaketitle

\section{Introduction}\label{sec:introduction}


%
%
%
%
\IEEEPARstart{P}hotos play a more important role in daily life nowadays. The popularization of the smartphone and the advance of image processing applications also make image editing easily. Thus, any ordinary person, who has little knowledge of image editing, can synthesize two images together by photo manipulation software~(Such as Photoshop) automatically under several easy steps~\cite{Cun:2018uj}: The composite image is synthesized by cutting region of the donor image and then paste it to the host image with several post-processing techniques, such as Gaussian Smoothing.
However, human eyes can easily identify the images are composited or not according to the color, texture or luminance differences between the splicing region and background. Meanwhile, these artifacts in the images degrade the visual quality of the whole image and make the image unrealistic. 
For these reasons, automatically image harmonization is an essential task in the digital image processing community. 
To address these problems, some previous works increase the quality of the synthesized image by hand-crafted features, such as color~\cite{Lalonde:2007wv} and texture~\cite{Sunkavalli:2010uv}. 
Nonetheless, these methods only focus on one particular kind of feature, causing unreliable results when the appearances are vastly different~\cite{Tsai:2017kv}. 
Recently, driven by deep learning based techniques, Zhu \textit{et al.}\cite{Zhu:2015tl} train a neural network to predict the realism of the composite image and optimize the color of the input image by freezing the parameters of the neural network, \cite{Tsai:2017kv} harmonize the images by an end-to-end neural network with semantic guidance.
However, their image harmonization method solves the non-convex model by multiple running~\cite{Zhu:2015tl} or rely on the power of learning directly~\cite{Tsai:2017kv}.

\begin{figure}[t]
  \centering
  \includegraphics[width=\columnwidth]{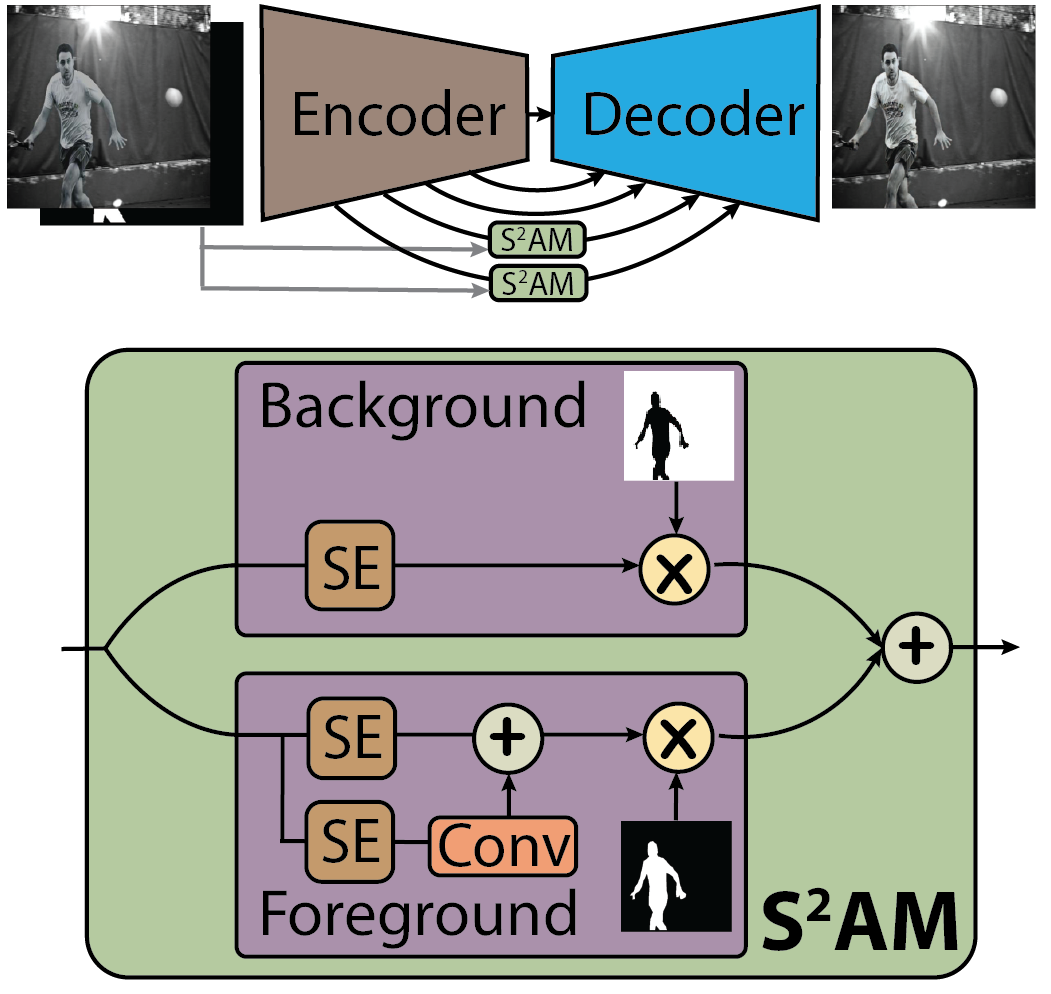}
  \caption{The basic idea of our method. In the task of image harmonization, we argue that the spliced image~(left) and the harmonized image~(right) share the same high-level features as well as the appearances in the non-spliced region. The difference mainly comes from the appearance in the spliced region. So we design the S$^{2}$AM module to learn these differences and make sure the consistency in the non-spliced region in low-level features. In detail, with the restrict of the hard-coded mask, we use several SE-Blocks~\cite{Hu:2017tf} to re-weight the encoded features and learn the spatial location differently. This module is inserted in the coarser levels features of the encoder-decoder structure since the main network ensures consistency in high-level features. More detail of each component in S$^2$AM can be found in the Section.~\ref{sec:method}.}
  \label{fig:teaser}
	\vspace{-1.5em}
\end{figure}

%

Different from the previous methods, our key observation is: \textit{For image harmonization task, the model needs to learn the dissimilarity in the appearance of the spliced region and ensure the consistency in the non-spliced regions and high-level features}.
As shown in the Fig.\ref{fig:teaser}, the spliced image and harmonized result only have differences in the appearance of the spliced region. This is because, from the viewpoint of image forgery detection, the artifacts in the composite images mainly come from the differences between various low-level features, such as camera internal features~\cite{Bondi:2017ek}, different compression levels in JPEG compression~\cite{Ye:2007tf} and noise levels~\cite{Pun:2016bp}. Consequently, we need to design an algorithm that pays closer attention to learning the appearance differences in a certain region. Meanwhile, the spliced image and the harmonized result describe the same story by analyzing the semantic information and the sense category. So, in our proposed method, the input spliced image and the harmonized result need to share the same high-level information.

To fit previous analysis, we proposed a novel framework for image harmonization task. 
Firstly, following the previous work on this task~\cite{Tsai:2017kv}, we regard the image harmonization as a supervised learning problem by synthesizing the composite images from some region of the natural image rather than the real copy-and-paste image. This alternative approach allows us to train the neural network in a supervised manner by considering the composite image and mask as input and natural image as the target. 
Then, we propose a novel attention module named Spatial-Separated Attention Module~(S$^{2}$AM) to learn the feature changes in the masked region and the others separately. Our S$^{2}$AM module uses several channel attention gates to weight features for different purposes with the help of the hard-coded mask and learns the changes in the specific masked region by a learning-able block. 
Moreover, we design the new framework of the image harmonization task. Our network base on the encoder-decoder~\cite{Isola:2016tp} by inserting the S$^{2}$AM module to low-level features in two different ways. Thus, the proposed S$^{2}$AM module learns the regional appearance changes in low-level features, and the encoder-decoder structure will ensure the consistency in high-level features as well. 
Finally, we make a big step for image harmonization without giving mask as well. As far as we know, our method is the first approach for harmonizing the composite image without mask by deep learning. This goal is achieved by replacing the hard-coded mask in S$^{2}$AM with the spatial attention block and we also use the additional supervision by measuring the loss between the spatial attention map and the ground truth mask.

We summarize the contributions as follows:
\begin{itemize}

\item[*] We propose a novel attentive module named Spatial-Separated Attention Module~(S$^{2}$AM) to learn the features in the foreground and background area by hard-coded masks individuality.

\item[*] We design two variants of the proposed S$^{2}$AM in Unet structure for image harmonization task. Our network fits the intention of the difference in the spliced area and the similarities between high-level features and the non-spliced region. Besides, our network is fully-convolutional which is suitable for real-world image harmonization tasks.

\item[*] We generalize our method to the image harmonization task without ground truth mask by the spatial attention module, attention loss and generative adversarial network. 

\item[*] Experiments show that the proposed S$^2$AM module gains better results than other attention mechanisms and our image harmonization method shows better performance than the state-of-the-art methods. 

\end{itemize}

\section{Related Works}
We do a brief review of the recent development of image harmonization and image-to-image transformation tasks. Importantly, we review the recently proposed attention mechanisms for the comparison of the proposed S$^2$AM.

\textbf{Image Harmonization}
Recently, because of the efficient and impressive results, the neural network based methods have drawn much attention in image editing. However, in image harmonization field, there are still limited research works to solve this problem. Zhu \textit{et al.}~\cite{Zhu:2015tl} propose a fine-tuned model for predicting the realism of the composite images. They also fix the parameters of the model and adjust the color of the masked region of the input image by optimizing the predicted realism score. It can also be regarded as an image harmonization task. \cite{Tsai:2016id} proposes a method for exchange the sky of images while this method only works for the sky. An end-to-end learning based approach of image harmonization has been proposed by \cite{Tsai:2017kv}, they trained an end-to-end neural network with the ground truth image and segmentation mask. However, their method just learns everything from the power of the neural network. Another related work is GP-GAN~\cite{Wu:2017wq}. It utilizes the generator in GAN to synthesize the low-resolution images, and then, the high-resolution output is synthesized by the input image and the low-resolution outputs using traditional Gaussian-Poisson Equation. Their method focuses more on blending the edge of the two different images rather than the style transform in the images. Thus, this method also changes the appearance of the background while the image harmonization only focus on the spliced region. More recently, video harmonization methods have been proposed by \cite{Huang:2018ti} and \cite{Lee:2019th}. Their methods harmonize the video and insert the video to another video using spatial and temporal information respectively.

\textbf{Attention Mechanisms} 
Attention mechanisms draw the attention of researchers because these modules allocate the most informative components according to analysis the features themselves. \textit{Squeeze-and-Excitation Network}~\cite{Woo:2018wr} insert a channel attention module in the network to make the network itself more focus on the essential features by weighting the channels with global feature analysis and \textit{Convolutional Block Attention Module}~\cite{Hu:2017tf} extend this method by adding a spatial attention module for weighting the features in spatial space. For choosing the feature kernel with different size, \textit{Selective Kernel Networks}~\cite{Li:2019vy} is proposed. As for learning the specific region in spatial space, \cite{Liu:2018ui} propose \textit{Partial Convolution} for focusing on the masked region and \cite{Yu:2018uw} design \textit{Contextual Attention} to borrow the background information to hole filling. \cite{Yu:2018hf} use \textit{Gated Convolution}, which can also be regarded as the spatial attention mechanism for each convolution block, to update the parameters of the network and mask~(or attention map ) softly. However, these methods are designed for image in-painting task specifically.
Another noticeable direction of attention mechanism is self-attention. Self-Attention aims to capture the long-range dependencies in the image context. \textit{Non-local Network}~\cite{Wang:2017ut} extend the self-attention~\cite{Vaswani:2017ul} as a form of non-local means~\cite{Buades:2005vp} to capture long-range dependencies for video classification and \cite{Chen:2018tu} extends this method in GAN-based structure for generating divergence samples. Most recently, \cite{Cao:2019wb} model the \textit{Non-Local Network} and \textit{Squeeze-and-Excitation Network} in a new form called \textit{Global Content Network} for various computer vision tasks. However, these attention mechanisms are focus on learning the representation on the high level computer vision task other than digital image processing with low-level features change.

\textbf{Image-to-Image Transformation} 
Image harmonization can also be regarded as an image-to-image transformation task which focuses on the changes the appearance between the input and target. Chen \textit{et al.}~\cite{Chen:2017ta} propose a neural network for fast image processing with the Dilated Convolution~\cite{Yu:2016uc} to capture a larger reception field without reducing the feature map. By the excellent performance of guided filter~\cite{He:2013je} in edge preserving, \cite{Gharbi:2017ks} and \cite{Chen:2016ck} model this filter by the neural network. Taking the benefit of \cite{Chen:2017ta} and \cite{He:2013je}, \cite{Wu:2018uk} propose the deep guided filter for fast image processing.  When the content of the image changes significantly, pix2pix~\cite{Isola:2016tp} train the network for alignment datasets by GAN~\cite{Goodfellow:2014wp} while CycleGAN~\cite{Zhu:2017ua} is proposed for unsupervised learning. By combining the image-to-image transformation task with the attention network, Another noticeable direction of unsupervised image-to-image transformation is attention-based methods. \cite{Chen:2018wl, Mejjati:2018wu} extend CycleGAN with the attention network as an additional block and learning the foreground specifically. These methods generate a global attention mask for image-to-image transformation for specific objects only.

\begin{figure*}[hbt]
	\centering
 	\includegraphics[width=\textwidth]{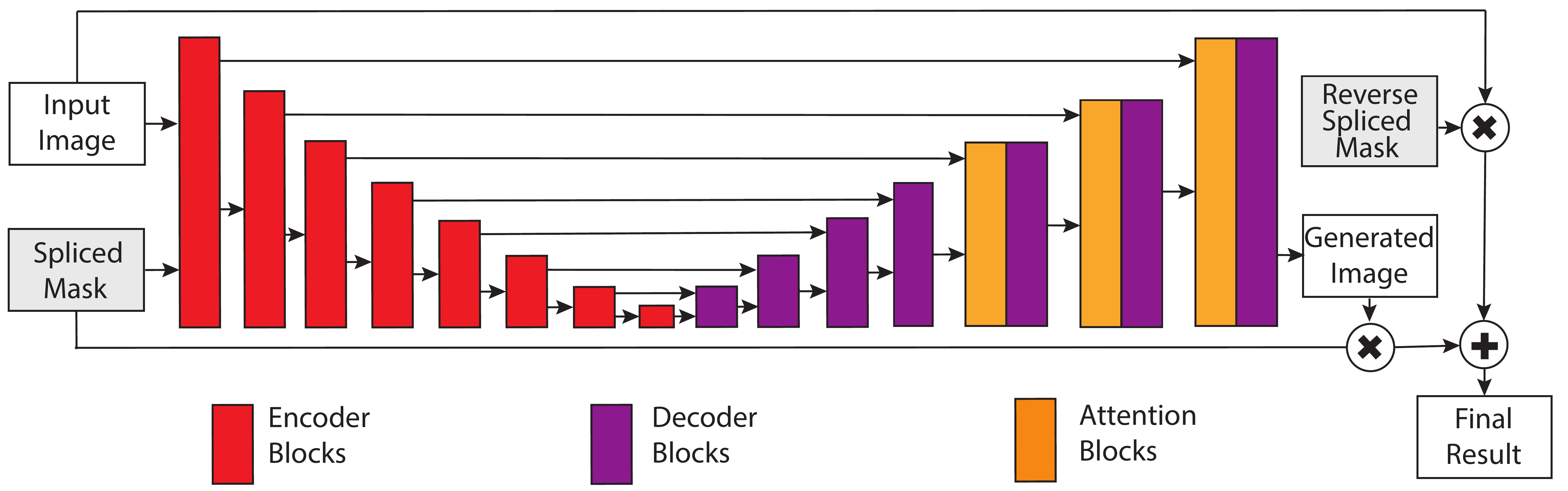}
 	\caption{The main framework of our network. }
 	\label{fig:main}
\end{figure*}

\section{Methods}
\label{sec:method}
Following the previous work~\cite{Tsai:2017kv} on image harmonization, we create synthesized image harmonization datasets by modifying the appearance of the specific region of the original natural image. Thus, image harmonization task can be modeled as a supervised learning task by feeding the synthesized composite image and the corresponding mask to the network and learning to harmonize the natural image.
Fig.~\ref{fig:main} shows an overview of our method for image harmonization. In detail, we use an encoder-decoder with skip-connection~(Unet in \cite{Isola:2016tp}) as the backbone structure for our task. 
Then, with the interpolating of the proposed attention block, our network can learn the spliced region and background region separately, causing better results than previous works. 
 In this section, we introduce the proposed Spatial-Separated Attention Module~(S$^{2}$AM) in Section.~\ref{sec:rasc} firstly. Then, for our method can learn the specific region automatically, we generalize our S$^{2}$AM module to image harmonization task without mask in Section.~\ref{sec:cam}. Next, we discuss the backbone network and the post-processing technique for image harmonization in Section.~\ref{sec:net}.  Finally, we discuss the loss functions in both tasks and synthesized datasets used in our framework in Section~\ref{sec:loss} and Section~\ref{sec:data}, respectively.

\begin{figure}[tb!]
\centering     
\subfigure[S$^{2}$AM]{
	\label{fig:rasc_module}
	\centering
	\includegraphics[height=0.7\columnwidth]{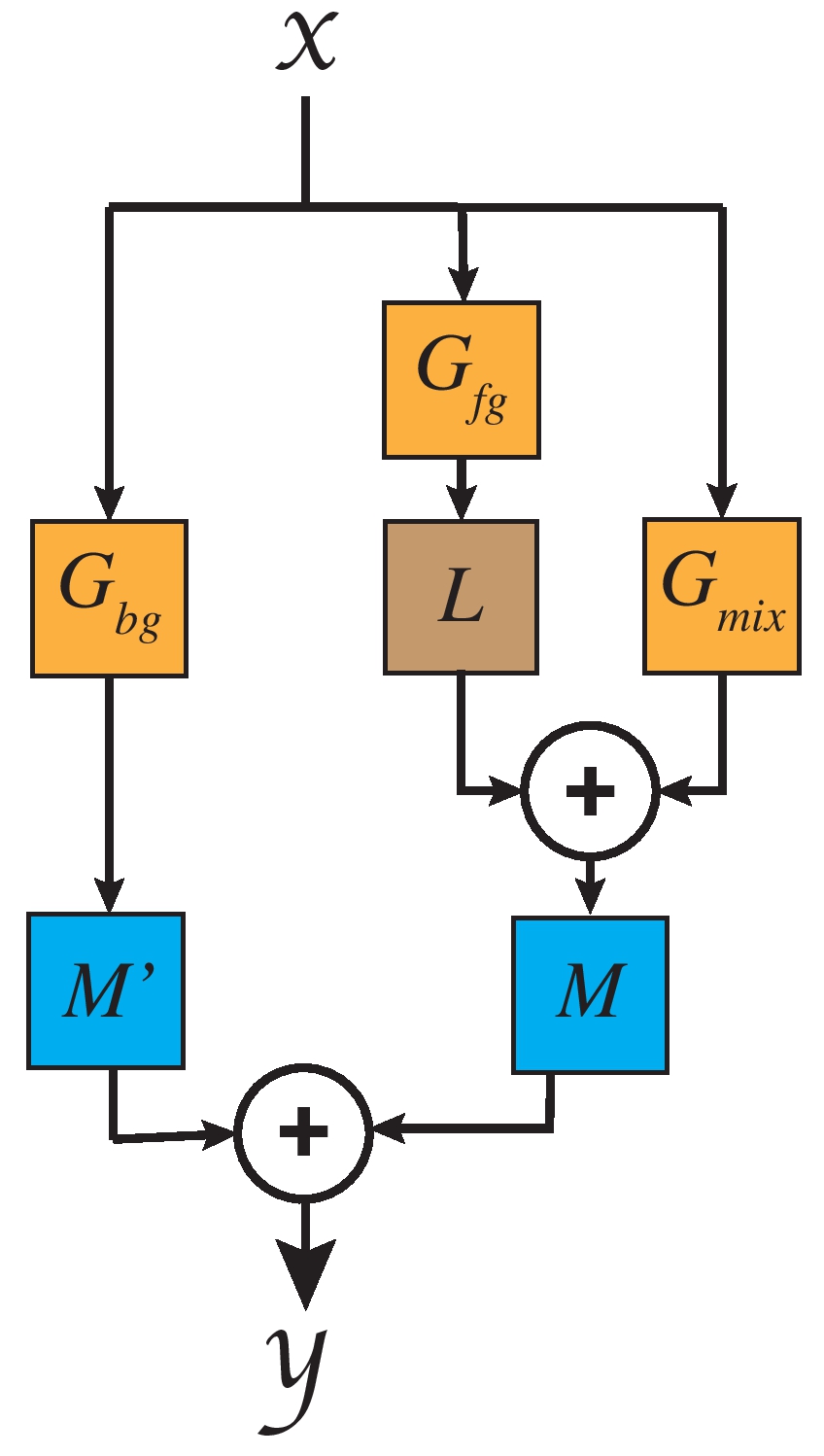}
}
\subfigure[S$^{2}$AM w/o hard-coded Mask]{
	\label{fig:global_attention}
	\centering
	\includegraphics[height=0.7\columnwidth]{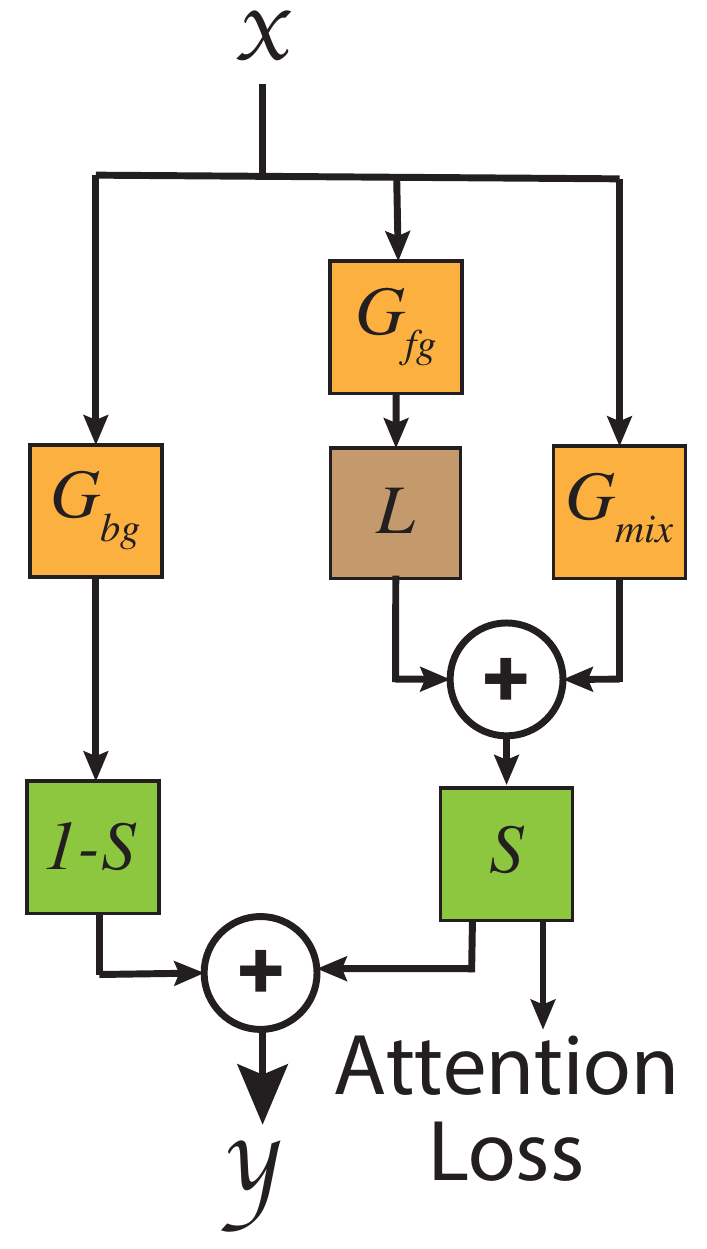}
}
\caption{The difference between the S$^{2}$AM and S$^{2}$AM without the hard-coded mask. We replace the original hard-coded mask by spatial attention block and attention loss from the ground truth mask. }
\vspace{-1em}
\label{fig:rasc}
\end{figure}

\subsection{Spatial-Separated Attention Module (S$^{2}$AM)}
\label{sec:rasc}
The basic idea behind our S$^{2}$AM module is straightforward: the traditional convolution filter in convolution neural network share the kernel parameters over the whole image, which benefits higher-level tasks to look for a given feature everywhere in the image, rather than in just a certain area. However, for the image-to-image transformation in the certain region, ~such as image harmonization and inpainting, it shows a relatively poorer performance~\cite{Liu:2018ui,ulyanov2018deep}. Thus, with the help of the hard-coded mask, we design the attention module attempting to learn the features individually in the spatial space.

As illustrated in Fig.~\ref{fig:rasc_module}, given an intermediate feature map $x\in\mathbb{R}^{C \times H \times W} $ and the binary mask $M \in \mathbb{R}^{1 \times H \times W}$ of the spliced region as the inputs~(~$C$ is the feature channel, $H$ and $W$ is the height and width of the feature map respectively~), our attention module utilizes three 1D channel attention gates $G_{bg}, G_{mix}, G_{fg}$ to weight the input features for different purposes with the help of hard-coded spliced region mask $M$ and non-spliced region mask $M'$. In detail, $G_{bg}$ aims to select the features which are necessary for image rebuild in the non-spliced region. $G_{fg}$ intends to weight the input features which have differences between the input and target. Moreover, we design the third channel attention module $G_{mix}$ to weight features which is unnecessary to change in the spliced region. Additionally, for the output features of $G_{fg}$, we add a learning-able block $L$ to learn the style transfer. So our RAB module between input features $x$ and output features $y$ can be formalized as:
\begin{equation}
  y = M\times[L(G_{fg}(x)) + G_{mix}(x)]+ (1-M) \times G_{bg}(x)
  \label{eq:rasc}
\end{equation}

Our S$^{2}$AM module is a general form on separated-learning the appearance changes in the specific region and others. So it is possible to design different structures of Channel Attention Modules $G_{bg}, G_{mix}, G_{fg}$ and learning-able block $L$ for the particular computer vision tasks. Here, inspired by the state-of-the-art attention modules, we design one instance of S$^{2}$AM for image harmonization as follows. 

\textbf{Channel Attention Modules~($G_{bg}, G_{mix}, G_{fg}$)}. With the help of hard-coded mask, the goal of the \textit{Channel Attention Module} is to select the related features channels from the input by applying 1D channel weighting for input features. Here, we modify the state-of-the-art channel attention module in \textit{Squeeze-and-Excitation Network}~\cite{Hu:2017tf} by adding a MaxPooling layer. This channel gate is also a part of the structure in \cite{Woo:2018wr}. As shown in Fig.~\ref{fig:global_attention}, we start from squeezing the input features $x\in\mathbb{R}^{C \times H \times W}$ to $x_{maxpooling}\in\mathbb{R}^{C \times 1 \times 1}$ and $x_{avgpooling}\in\mathbb{R}^{C \times 1 \times 1}$ by the \textit{Global Max Pooling} and \textit{Global Average Pooling} operators, separately. Then, we concatenate $x_{avgpooling}$ and $x_{maxpooling}$ in channel axis and learn the channel weighting by two fully-connected layers and a Sigmoid layer. Following previous work~\cite{Hu:2017tf,Woo:2018wr}, we reduce the size of the input feature to $\frac{C}{16}$ in the first Linear layer and increase the length to $C$ in the second and a Sigmoid layer is added at the end of Linear layer to ensure the range of scale factor under $[0,1]$. The attentive modules show great success in image classification~\cite{Hu:2017tf} and object detection tasks~\cite{Woo:2018wr} with the structure of ResNet~\cite{He:2015tt} or Inception~\cite{Szegedy:2014tb}. As far as we know, we are the first to insert the attentive structure for the module with the hard-coded mask.

\textbf{Learning-able Block ($L$)} As shown in Fig.~\ref{fig:rasc_module}, we aim to learn the appearance changes in the spliced regions by Learning-able Block additionally. We use $ 3 \times 3 $ convolutional layers with two CONV-BN-ELU blocks. Similar to channel attention module, we double the channel size in the first CONV-BN-ELU block and reduce the channel size in the second to expend the learning capacity.  Notice that, we do not reduce the feature size in the learning-able blocks for preserving the feature details. 

\textbf{Mask ($M,M'$)} Spliced region mask $M$ and non-spliced region mask $M'$ are the necessary parts for our S$^{2}$AM module because the channel attention modules are guided by these hard-coded masks. Furthermore, they provide a strong prior knowledge to the neural network. We also multiply a uniform $7\times7$ Gaussian Kernel to the mask $M$ and the reverse mask $M'$ respectively. This simple operation allows the neural network to generate the boundary by the closest object for the hard-coded masks are not always accurate. We discuss the effect of Gaussian Kernel in our attention module in the Section.~\ref{sec:astudy}

\begin{figure}[tb!]
\centering     
\subfigure[Channel Attention Module]{
	\label{fig:global_attention}
	\centering
	\includegraphics[width=\columnwidth]{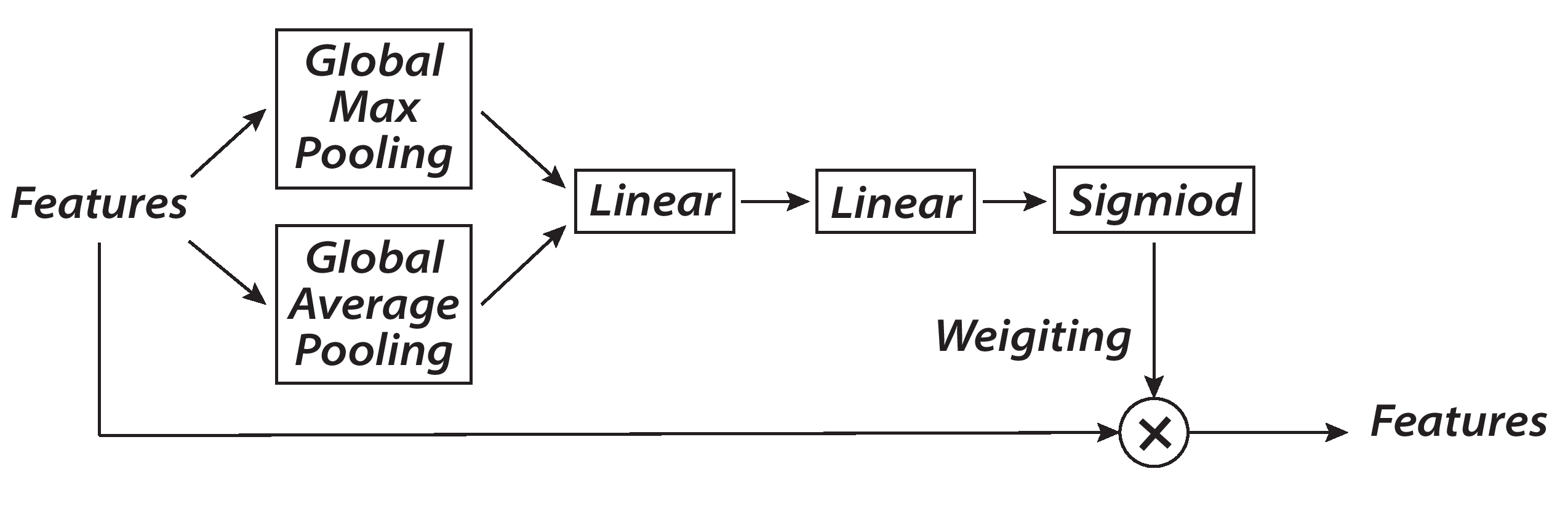}
}
\subfigure[Spatial Attention Module]{
	\label{fig:spatial_attention}
	\centering
	\includegraphics[width=\columnwidth]{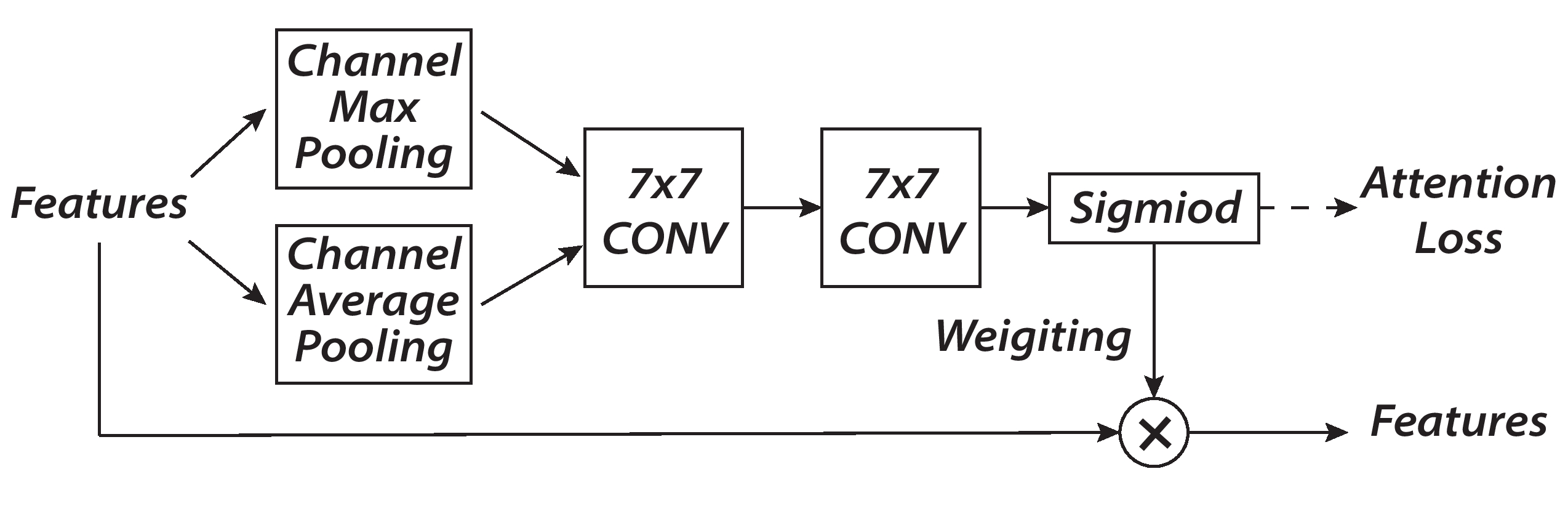}
}
\caption{The detail of the \textit{Channel Attention Module} and \textit{Spatial Attention Module} in the proposed S$^{2}$AM. }
\label{fig:rasc}
\end{figure}

\subsection{Generalizing S$^{2}$AM to Image Harmonization without Mask}
\label{sec:cam}

%
%
%
%

We generalize the S$^{2}$AM module for image harmonization without the hard-coded mask as input. Thus, we first replace the original hard-coded mask with the spatial attention block. We design the spatial attention module with the similar structure of the previous introduced \textit{Channel Attention Modules}. For the \textit{Spatial Attention Module} aims to learn the spatial position separately, we use the \textit{Max Channel Pooling} and \textit{Average Channel Pooling} to extract the maximum responses and medium responses from all feature channels for each pixel. Then, unlike the channel attention module which used fully-connected layer for channel weighting. It is memory-consuming \textit{to} capture global features. Thus, we use two Convolutional layers with the kernel size $7\times7$ as alternative. The first Convolutional layer broadcast the features to 16 layers while the second squeeze the channel map to fit the output of Sigmoid. Then, a Sigmoid layer is added at the end of the module to restrict the response of features weighting in $[0,1]$. Notice that, there is only one \textit{Spatial Attention Module} in our S$^{2}$AM module because we attempt to force the spatial attention map in two folds without overlap. The reverse mask $M'$ is calculated by $M' = 1-M$. We show the structure of the spatial attention module in Fig.~\ref{fig:spatial_attention}.

Moreover, we add a loss between the output of the sigmoid layer and the ground truth mask. This attention loss guide the focus of spatial attention module for the features from $7\times7$ Convolutional block is not always accurate. We review the choice of attention loss and the influence of attention loss in Section.\ref{sec:loss} and Section.~\ref{sec:comw}.

%

\subsection{Network Structure}
\label{sec:net}
As the previous analysis, we model the image harmonization task as an image-to-image transformation task. Differently, our task needs to: (1) ensuring the consistence in high level features. (2) taking more attention to the low-level appearance changes. Thus, we use the encoder-decoder structure as the backbone network to ensure the consistency in high-level features and S$^{2}$AM for learning appearance changes carefully. We introduce the network structure in our approach below.

\textbf{Network Backbone} Different from the encoder-decoder structure in \textit{Deep Image Harmonization (DIH)}~\cite{Tsai:2017kv}, the state-of-the-art work on image harmonization, we use Unet~\cite{Isola:2016tp, Ronneberger:2015vw} as our backbone network. Unet owns more high-level filters and treats the global features in the fully convolutional manner while DIH uses Linear Layer which will hugely increase the parameters of networks and limit the input size. 
Another difference between these two backbones is that DIH use ELU for non-linear activation while Unet chooses RELU/Leaky RELU. 
As for the operation between the skipped encoder features and decoder features, we concatenate the features from encoder rather than add or mix them. The experiment shows that the new baseline network improves the results of harmonization task to a large extent.

\begin{figure}[htb]
\centering     
\subfigure[S$^{2}$ASC]{\label{fig:rasc_v1}
\centering\includegraphics[width=0.45\columnwidth]{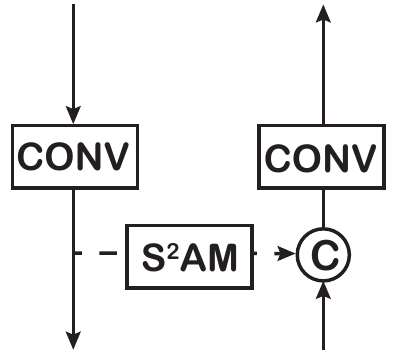}}
\subfigure[S$^{2}$AD]{\label{fig:rasc_v2}
\centering\includegraphics[width=0.45\columnwidth]{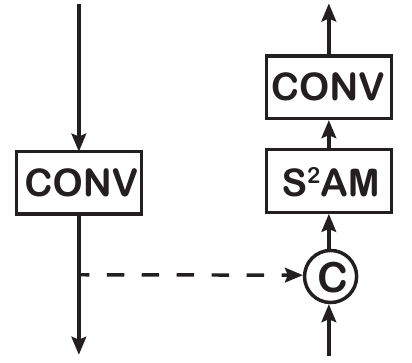}}
\caption{The positions of the proposed module.}
\label{fig:rascnetwork}
\end{figure}

\textbf{The Role of S$^{2}$AM} The first choice of interpolating S$^2$AM to Unet is based on a general observation: The original skip-connections preserve the encoded low-level features in the original image and these features are concatenated to the low-level features of the decoder. The concatenation of the low-level features from different appearances will mislead the learning process to make a good choice. Thus, we replace the original skip-connection in the encoder-decoder by the proposed attention module. Our attention module will filter the features in the encoder and learn the appearance changes in the spliced region and non-spliced region individually. We call this type of connection \textit{Spatial-Separated Attentive Skip-Connection}~(\textit{\textbf{S$^{2}$ASC}}).
Moreover, we also interpolate the proposed module after the concatenation of the shadow features from the encoder and the decoded features from high-level network~(as the network structure in Fig.~\ref{fig:main}). This kind of choice will use more features (features from encoder and decoded features from high-level) for learning a better channel attention decision to learn separately. Thus, the appearance changes will also contain the influence of the global features. We name this type of structure \textit{Spatial-Separated Attentive Decoder}~(\textbf{\textit{S$^{2}$AD}}). We show the differences between S$^{2}$ASC and S$^{2}$AD in Fig.~\ref{fig:rascnetwork}.

Notice that, the proposed S$^{2}$ASC and S$^{2}$AD module only replace the original three levels of the structure in the unet for ensuring a similar high-level feature (as in Fig.~\ref{fig:main}). We also verify this assumption in Section.~\ref{sec:astudy}.

\textbf{Post-Processing~(pp)} As for image harmonization task, the background region should stay unchanged when we generate the final output. Thus, in the inference stage, we mask the harmonized region by the original ground truth mask and replace the others with the original image. 
For image harmonization without mask, we train the post-processing altogether. The final output $I_{final}$ is calculated by the coarser attention mask $M_{2s}$ and input image $I$ of our network as:
\begin{equation}
	I_{final} = I_{g} \otimes Up(M_{2s}) + I \otimes(1-Up(M_{2s})) 
\end{equation}
where $\otimes$ represent the pixel-wise multiplication and $I_g$ is the output of the network. $Up(\cdot)$ means the Upsample layer for the attention mask 2s is the $1/2$ size of the original image. Notice that we learn the whole network in image harmonization without mask task while we use this technique as post-processing in image harmonization task. This is because the predicted mask is not always accurate and the additional loss in image harmonization without mask will learn to make the overall image look realistic rather than the attention mask region only.

\subsection{Loss Function}
\label{sec:loss}
We design different loss functions for image harmonization and image harmonization without mask since the latter task have  weaker prior information.

\subsubsection{Loss Function for Image Harmonization}
Following our baseline method~\cite{Tsai:2017kv}, we use the $L_2$-norm criterion as the main criterion in the pixel domain. Besides, \textit{Perception Loss}~\cite{Johnson:2016wm} and \textit{Adversarial Loss}~\cite{Goodfellow:2014wp}~(~pix2pix method in Section.~\ref{sec:exp}~) are added individuality as the additional loss to constrict the accuracy of results in the feature domain and distribution. These two criteria have shown great success in various image generation tasks. However, we do not observe a significant improvement in our task which focuses on low-level feature processing. Similar experiment results have been observed in image processing~\cite{Chen:2017ta}.

\subsubsection{Loss Function for Image Harmonization w/o Mask}
Different from previous image harmonization task, we use multiple criteria for Image harmonization without the mask. 
Firstly, as the same loss function in image harmonization, we set the pixel-level criterion by $L_2$ norm for the harmonized image should have the same pixel values with the target in the corresponding spatial position. We mark the output image as $I_{predict}$ and the target image as $I$. The pixel level loss function can be written as:
\begin{equation}
	L_{pixel} = || I_{predict} - I||_2
\end{equation}

Although our spatial attention module can learn the foreground features and background features individually, it is still a challenging task when there is no ground truth mask as input. However, for the ground truth spliced mask is easy to get by the synthesized dataset, we add the attention loss in the spatial attention module of the S$^{2}$AM module for guiding the image synthesis. 
In specific, we resize the original ground truth mask to constrict the output of the sigmoid in every level of spatial attention mask. It is obvious that the gated attention can be considered as a supervised binary classification task, we use the Binary Cross Entropy as the criterion firstly. Then, every pixel in the image has spatial connections. Thus, $L_2$ Loss is also a satisfying option. We evaluate all the choices of the loss function in Section~\ref{sec:comw}. Finally, we choice $L_2$ loss as the attention loss in each spatial attention block:
\begin{equation}
	L_{att} = \sum_{i=1}^{3}|| M_{i}^{'} - M_{i}||_2
\end{equation}
$M_{i}^{'}$ is the $i$-th level attention mask of S$^{2}$AM while $M_{i}$ is the corresponding ground truth spliced mask.

Moreover, we use the adversarial loss~\cite{Isola:2016tp} for removing the artifacts in the predicted image. Adversarial loss has shown much success in many image-to-image transformation tasks and image generation tasks. Different from original works in Pix2pix~\cite{Isola:2016tp}, we use the Least-Square GAN as the adversarial loss following the work of CycleGAN~\cite{Zhu:2017ua}. For image harmonization task, we regard the Unet with S$^{2}$AD is the generator of GAN and we use the same Patch-based discriminator structure of pix2pix~\cite{Isola:2016tp}.

Then, the total loss for image harmonization without mask can be written as:
\begin{equation}
  L = \alpha  L_{attention} + \beta L_{pixel} + L_{gan}
\label{eq:att}
\end{equation}
We experimentally set the $\alpha = 90$ and $\beta = 100$ in all experiments. We also evaluate the different settings of hyper-parameters $\alpha, \beta$ in Section.~\ref{sec:comw}. 

\subsection{Data Acquisition}
\label{sec:data}
Since there is no publicly available dataset for image harmonization, we build the below two datasets to compare for different reasons.

\begin{figure}[ht!]
 \centering
  \includegraphics[width=\columnwidth]{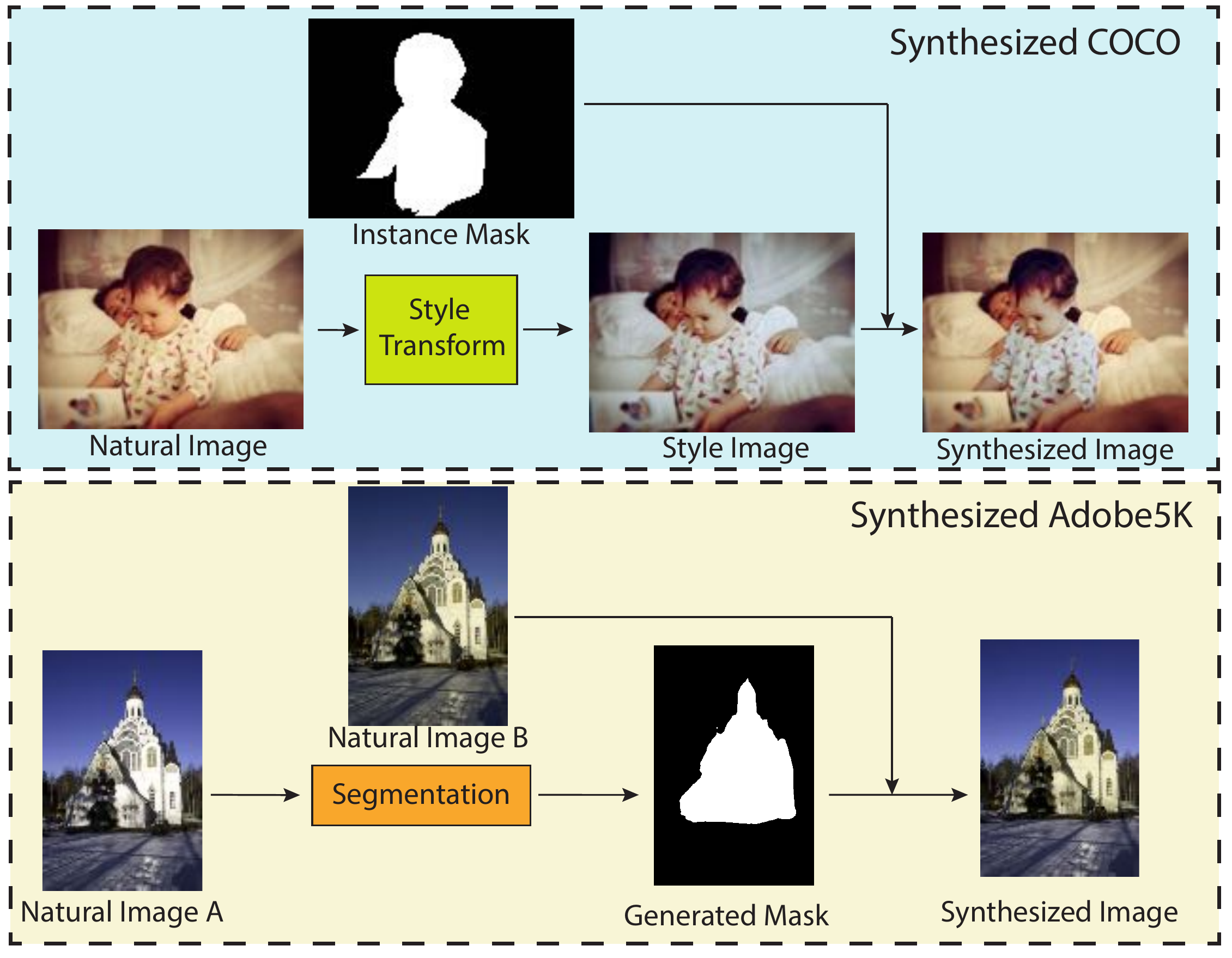}

  \caption{The workflow of building synthesized datasets.}
  \label{fig:dataset}

\end{figure}

\textbf{Synthesized COCO~(S-COCO)}
To test the robustness of approaches over the tampered images with ground truth instance mask, we build the S-COCO dataset. 
COCO~\cite{Lin:2014vm} contains various natural images with ground truth instance mask annotated by the human. 
Different from DIH~\cite{Tsai:2017kv}, we tamper the images by the state-of-the-art style transfer methods~\cite{Wu:2018uk}. 
As shown in Fig.~\ref{fig:dataset}, firstly, we temper the original natural image by the pre-trained \textit{auto-ps} model. 
Then, for further simulating the copy-and-paste image in the real world, we randomly enhance the appearance of the transformed region with brightness, color and contrast. 
Finally, we crop the tampered image by the ground truth instance mask and paste the cropped region to the corresponding position of the natural image.
 We use the Train2017 and Val2017 split in COCO for training and testing, respectively. 
 We only select the tampered image with the spliced region larger than \SI{15}{\percent} of the whole image since smaller regions may harm the consistency of the prediction. 
 Overall, there are about 43k and 1.7k images for training and testing respectively.

\textbf{Synthesized Adobe5K~(S-Adobe5K)}
We evaluate the robustness of the image harmonization methods on inaccurate ground truth mask and manually copy-and-paste under Adobe5K dataset~\cite{Bychkovsky:2011wf}. Adobe5K contains 5k raw photos, and each image is dedicated to photo adjustment by five experts using Adobe Lightroom software. So each image has five different natural styles. To get the ground truth image spliced binary mask, we segment image from one specific style by the pre-trained state-of-the-art semantic segmentation model, DeepLab-V3~\cite{Chen:2017we}. We generate the binary mask by each class of segmentation results. Finally, the samples are created by pasting the mask region from other different styles to the original image. We split the original dataset for learning randomly and filter out the smaller region as S-COCO. There are 36k images for training and 2k images for testing in S-Adobe5K. 

The overall progress is shown in Fig~\ref{fig:dataset}. We will public these two synthesized datasets for further research.

\section{Experiments}
\label{sec:exp}
In this section, we conduct experiments on our methods with state-of-the-art methods on image harmonization. Firstly, we illustrate all the training/testing details of our methods and the baselines for comparisons in Section.~\ref{sec:imp} and Section.~\ref{sec:baseline} respectively. Then, for the comparison, we evaluate our method for image harmonization and image harmonization without the mask on several datasets in Section.~\ref{sec:com} and Section.~\ref{sec:comw}. Finally, a detailed analysis of the proposed S$^{2}$AM is reported in Section.~\ref{sec:attention}.

\subsection{Implementation detail} 
\label{sec:imp}
All the models are trained with PyTorch~\cite{paszke2017automatic} v1.0 and CUDA v9.0. We train the image with the resolution of $ 256 \times 256 $ and run the model \SI{80} and \SI{60} epochs for converging on S-COCO and S-Adobe5K, respectively. All the optimizers are Adam~\cite{Kingma:2014us} with the learning rate of \SI{0.001}, and there is no learning rate adjustment schedule for a fair comparison. For testing, our model runs at 0.012 seconds pre-image on a single NVIDIA 1080 GPU with a resolution of $ 256 \times 256 $.
Since the synthesized datasets have the ground truth target, we evaluate our approach with other methods on multiple popular numerical criteria, such as Mean Square Error (MSE), Structural Similarity (SSIM) and Peak Signal-to-Noise Ratio (PSNR). All the results in the synthesized datasets are the mean of the $1.7$k and $2.4$k test images in S-COCO and S-Adobe5K, separately. Moreover, following the metrics in previous works~\cite{Tsai:2017kv,Wu:2017wq}, a pre-trained realism prediction model~\cite{Zhu:2015tl} is used to evaluate all the results from the viewpoint of the neural network. Finally, We evaluate the state-of-the-art methods on the real composite images which are collected from the real world with a user study.

\subsection{Baselines}
\label{sec:baseline}

We list all the image harmonization baselines for comparison on the synthesized datasets as follows: 
\begin{figure*}[th!]
\centering     
\subfigure{\centering\includegraphics[width=0.21\columnwidth]{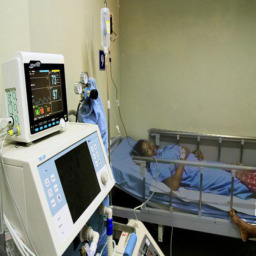}}
\subfigure{\centering\includegraphics[width=0.21\columnwidth]{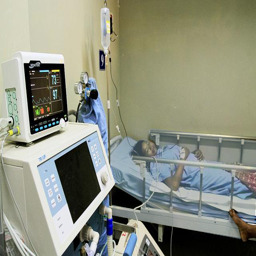}}
\subfigure{\centering\includegraphics[width=0.21\columnwidth]{figures/figurecoco/dih256seg_1448}}
\subfigure{\centering\includegraphics[width=0.21\columnwidth]{figures/figurecoco/rascseg_1448}}
\subfigure{\centering\includegraphics[width=0.21\columnwidth]{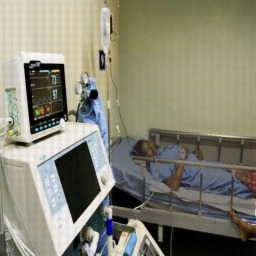}}
\subfigure{\centering\includegraphics[width=0.21\columnwidth]{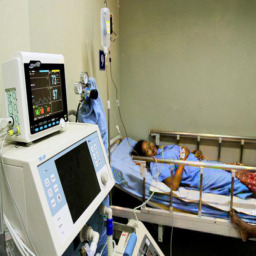}}
\subfigure{\centering\includegraphics[width=0.21\columnwidth]{figures/figurecoco/unet_1448}}
\subfigure{\centering\includegraphics[width=0.21\columnwidth]{figures/figurecoco/radhnv3xgaussian_1448}}
\subfigure{\centering\includegraphics[width=0.21\columnwidth]{figures/figurecoco/target_1448}}

\par\bigskip\vspace*{-2em}
\subfigure{\centering\includegraphics[width=0.21\columnwidth]{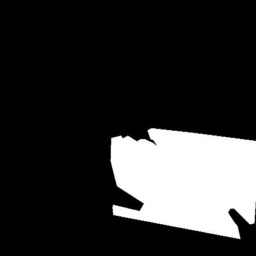}}
\subfigure{\centering\includegraphics[width=0.21\columnwidth]{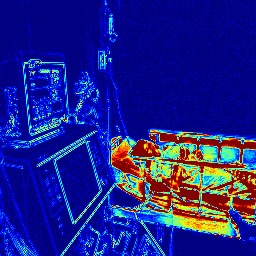}}
\subfigure{\centering\includegraphics[width=0.21\columnwidth]{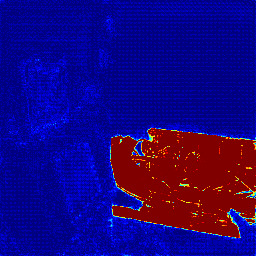}}
\subfigure{\centering\includegraphics[width=0.21\columnwidth]{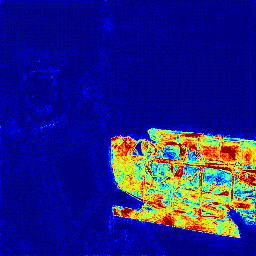}}
\subfigure{\centering\includegraphics[width=0.21\columnwidth]{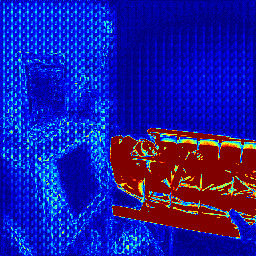}}
\subfigure{\centering\includegraphics[width=0.21\columnwidth]{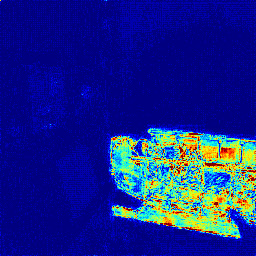}}
\subfigure{\centering\includegraphics[width=0.21\columnwidth]{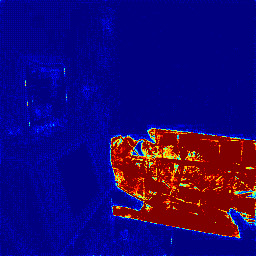}}
\subfigure{\centering\includegraphics[width=0.21\columnwidth]{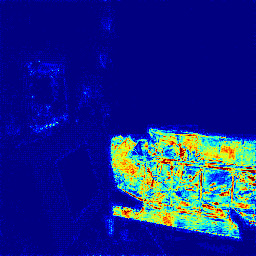}}
\subfigure{\centering\includegraphics[width=0.21\columnwidth]{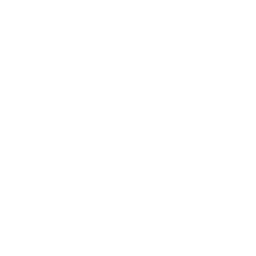}}

\par\bigskip\vspace*{-2em}
\subfigure{\centering\includegraphics[width=0.21\columnwidth]{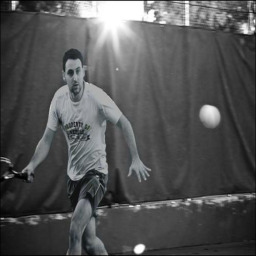}}
\subfigure{\centering\includegraphics[width=0.21\columnwidth]{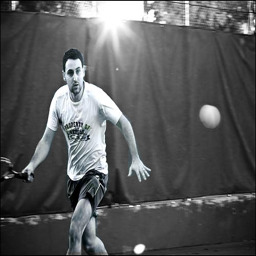}}
\subfigure{\centering\includegraphics[width=0.21\columnwidth]{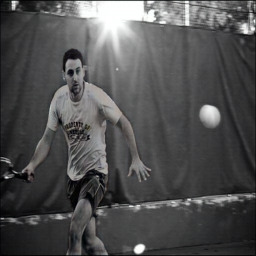}}
\subfigure{\centering\includegraphics[width=0.21\columnwidth]{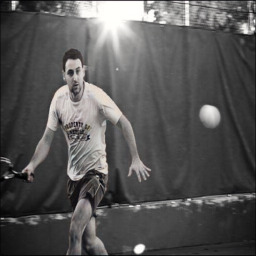}}
\subfigure{\centering\includegraphics[width=0.21\columnwidth]{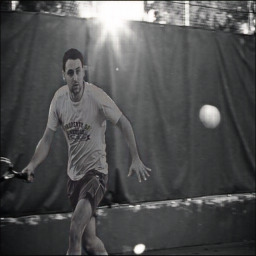}}
\subfigure{\centering\includegraphics[width=0.21\columnwidth]{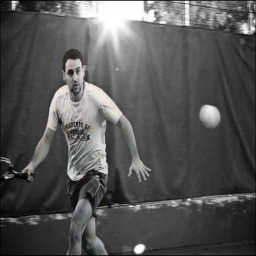}}
\subfigure{\centering\includegraphics[width=0.21\columnwidth]{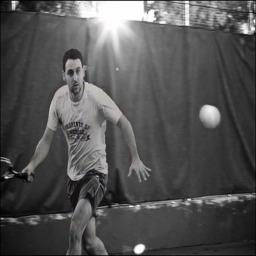}}
\subfigure{\centering\includegraphics[width=0.21\columnwidth]{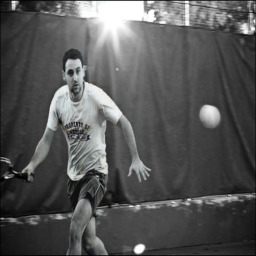}}
\subfigure{\centering\includegraphics[width=0.21\columnwidth]{figures/figurecoco/target_781}}

\par\bigskip\vspace*{-2em}
\subfigure{\centering\includegraphics[width=0.21\columnwidth]{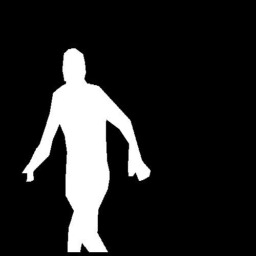}}
\subfigure{\centering\includegraphics[width=0.21\columnwidth]{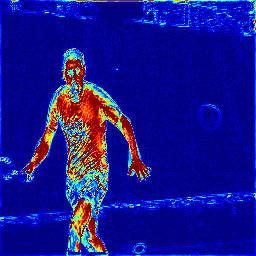}}
\subfigure{\centering\includegraphics[width=0.21\columnwidth]{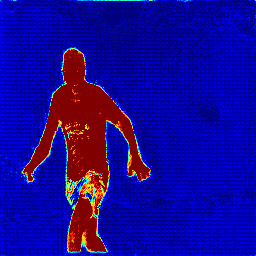}}
\subfigure{\centering\includegraphics[width=0.21\columnwidth]{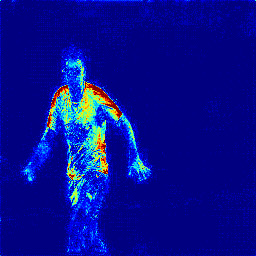}}
\subfigure{\centering\includegraphics[width=0.21\columnwidth]{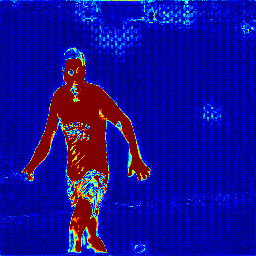}}
\subfigure{\centering\includegraphics[width=0.21\columnwidth]{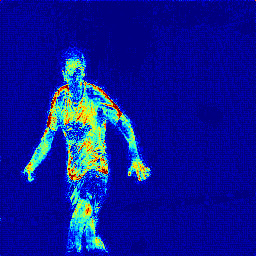}}
\subfigure{\centering\includegraphics[width=0.21\columnwidth]{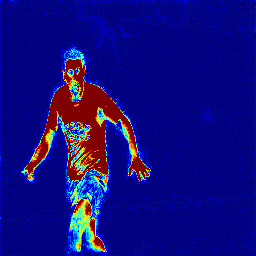}}
\subfigure{\centering\includegraphics[width=0.21\columnwidth]{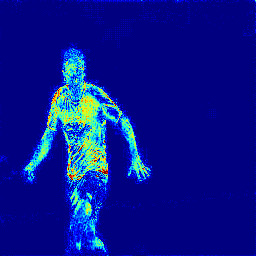}}
\subfigure{\centering\includegraphics[width=0.21\columnwidth]{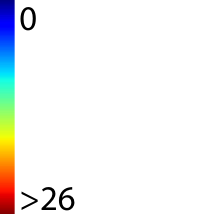}}
\par\bigskip\vspace*{-2em}

\subfigure{\centering\includegraphics[width=0.21\columnwidth]{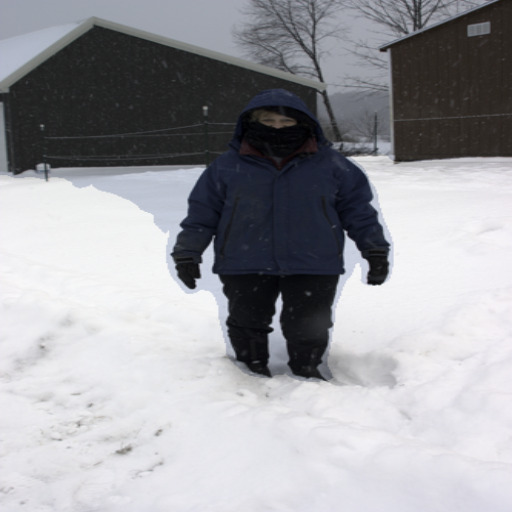}}
\subfigure{\centering\includegraphics[width=0.21\columnwidth]{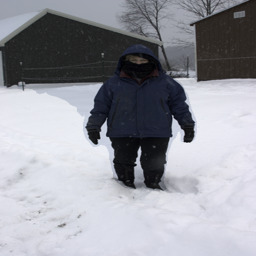}}
\subfigure{\centering\includegraphics[width=0.21\columnwidth]{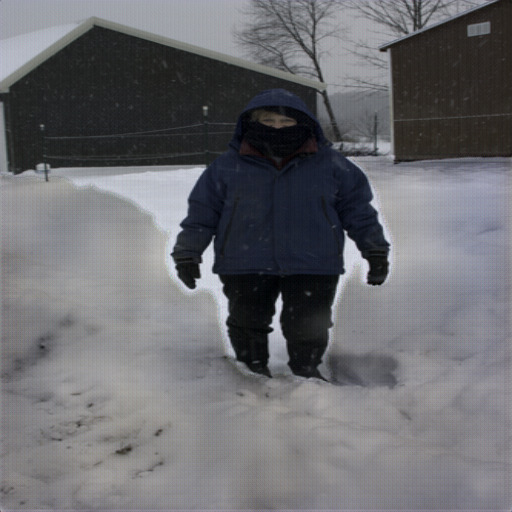}}
\subfigure{\centering\includegraphics[width=0.21\columnwidth]{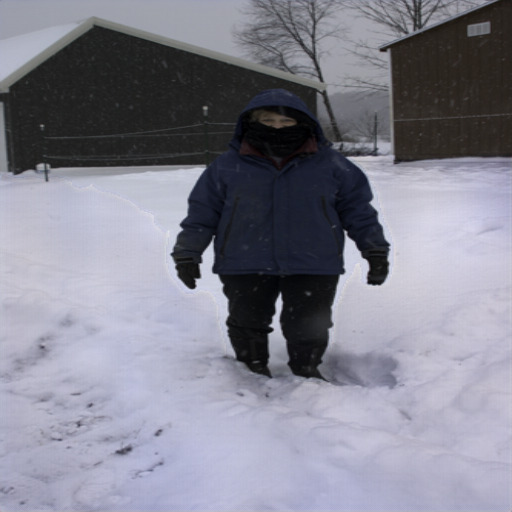}}
\subfigure{\centering\includegraphics[width=0.21\columnwidth]{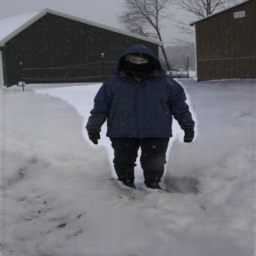}}
\subfigure{\centering\includegraphics[width=0.21\columnwidth]{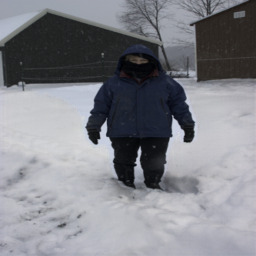}}
\subfigure{\centering\includegraphics[width=0.21\columnwidth]{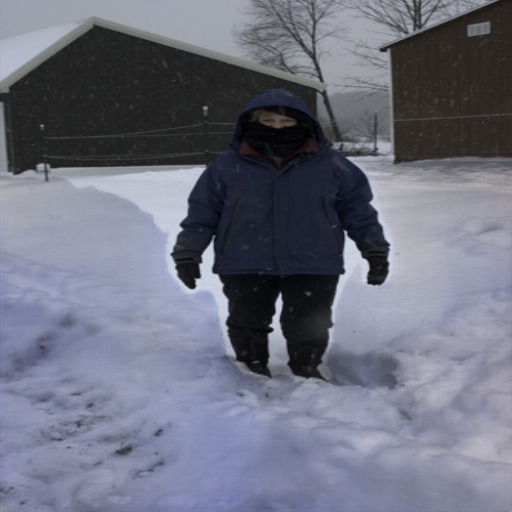}}
\subfigure{\centering\includegraphics[width=0.21\columnwidth]{figures/figurecoco/radhnv3xgaussian_668}}
\subfigure{\centering\includegraphics[width=0.21\columnwidth]{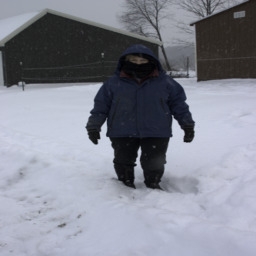}}

\par\bigskip\vspace*{-2em}

\subfigure{\centering\includegraphics[width=0.21\columnwidth]{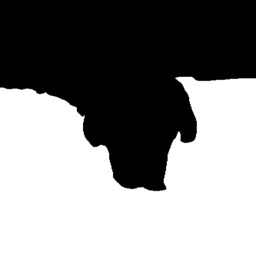}}
\subfigure{\centering\includegraphics[width=0.21\columnwidth]{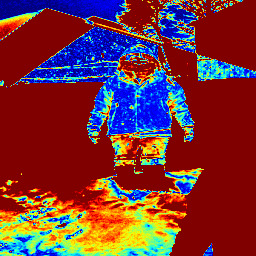}}
\subfigure{\centering\includegraphics[width=0.21\columnwidth]{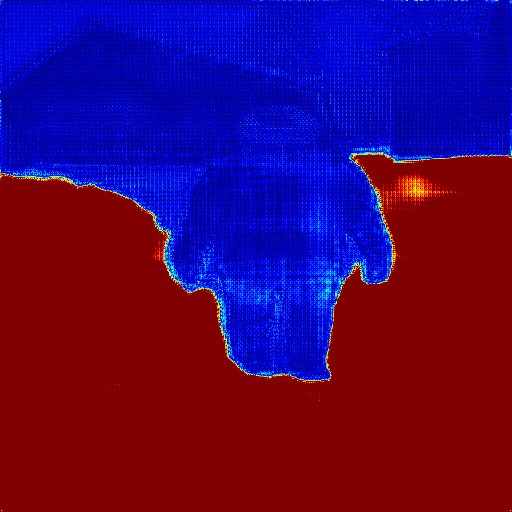}}
\subfigure{\centering\includegraphics[width=0.21\columnwidth]{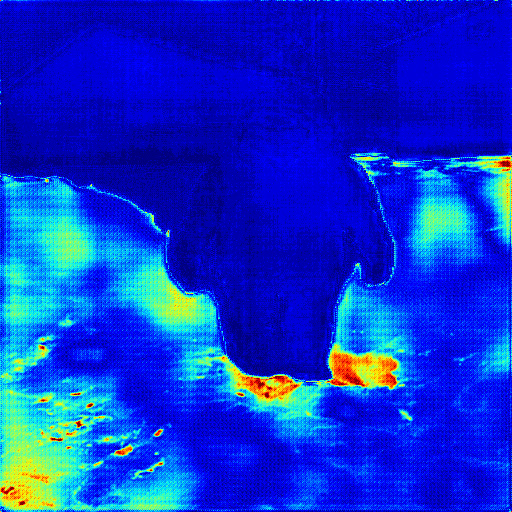}}
\subfigure{\centering\includegraphics[width=0.21\columnwidth]{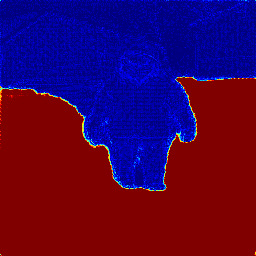}}
\subfigure{\centering\includegraphics[width=0.21\columnwidth]{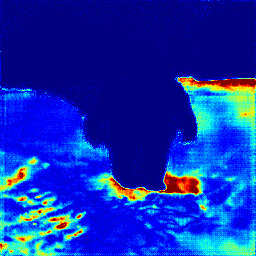}}
\subfigure{\centering\includegraphics[width=0.21\columnwidth]{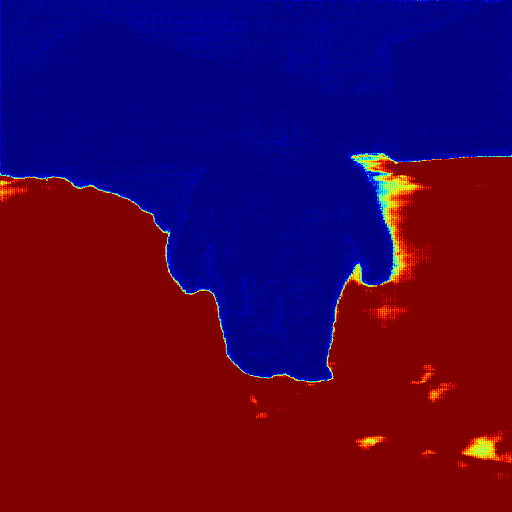}}
\subfigure{\centering\includegraphics[width=0.21\columnwidth]{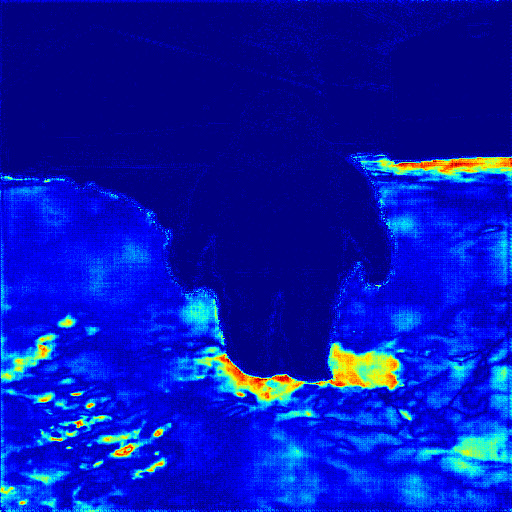}}
\subfigure{\centering\includegraphics[width=0.21\columnwidth]{figures/placeholder}}

\par\bigskip\vspace*{-2em}
\subfigure{\centering\includegraphics[width=0.21\columnwidth]{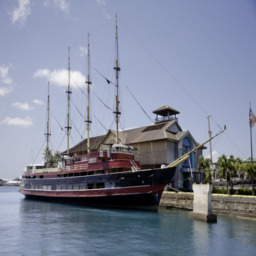}}
\subfigure{\centering\includegraphics[width=0.21\columnwidth]{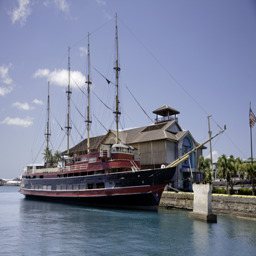}}
\subfigure{\centering\includegraphics[width=0.21\columnwidth]{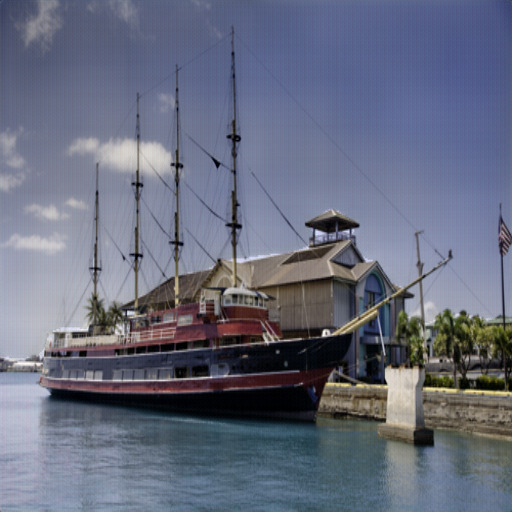}}
\subfigure{\centering\includegraphics[width=0.21\columnwidth]{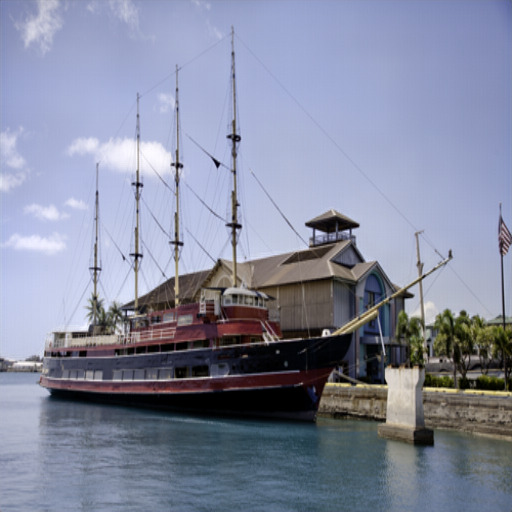}}
\subfigure{\centering\includegraphics[width=0.21\columnwidth]{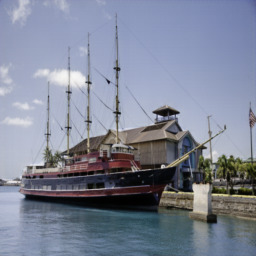}}
\subfigure{\centering\includegraphics[width=0.21\columnwidth]{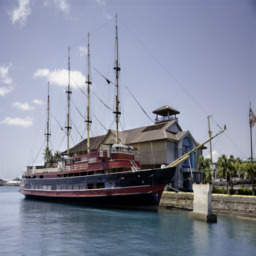}}
\subfigure{\centering\includegraphics[width=0.21\columnwidth]{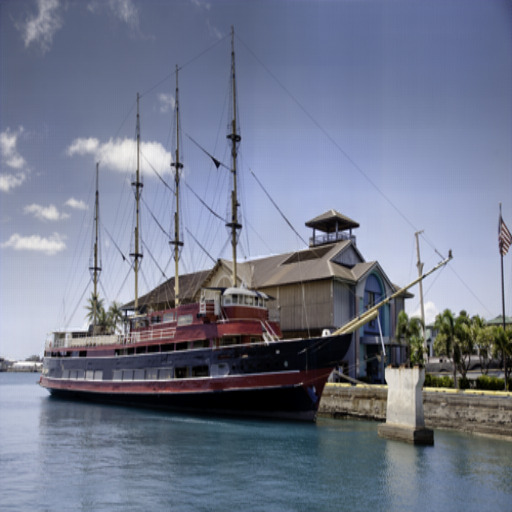}}
\subfigure{\centering\includegraphics[width=0.21\columnwidth]{figures/figurecoco/radhnv3xgaussian_712}}
\subfigure{\centering\includegraphics[width=0.21\columnwidth]{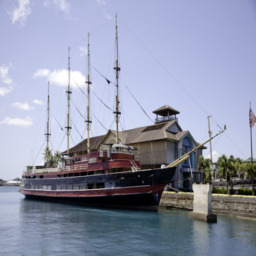}}

\par\bigskip\vspace*{-2em}

\addtocounter{subfigure}{-63}
\subfigure[Input/Mask]{\centering\includegraphics[width=0.21\columnwidth]{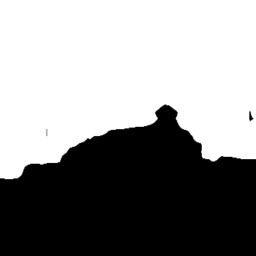}}
\subfigure[R-CNN\cite{Chen:2017ta}]{\centering\includegraphics[width=0.21\columnwidth]{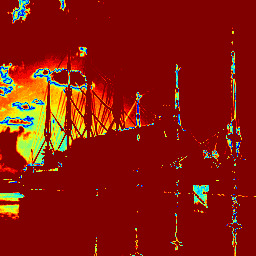}}
\subfigure[DIH\cite{Tsai:2017kv}]{\centering\includegraphics[width=0.21\columnwidth]{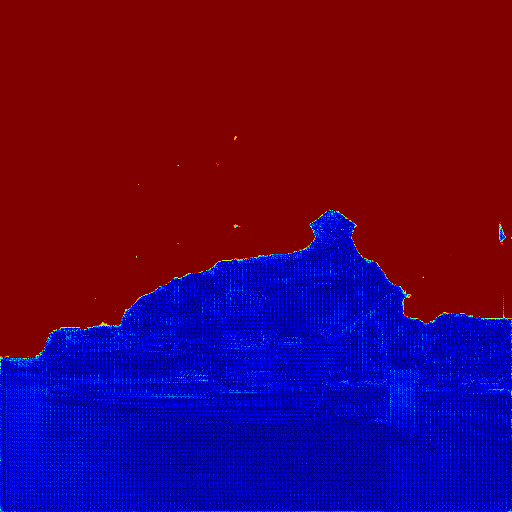}}
\subfigure[DIH+]{\centering\includegraphics[width=0.21\columnwidth]{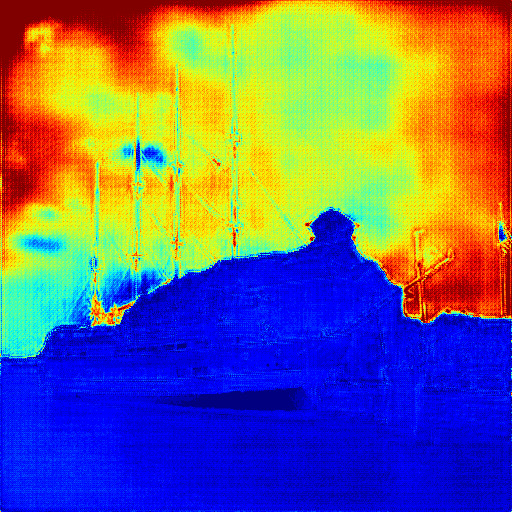}}
\subfigure[pix2pix]{\centering\includegraphics[width=0.21\columnwidth]{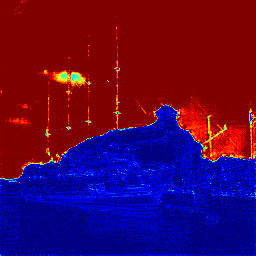}}
\subfigure[pix2pix+]{\centering\includegraphics[width=0.21\columnwidth]{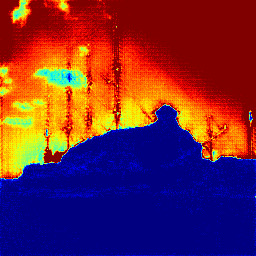}}
\subfigure[UNET\cite{Isola:2016tp}]{\centering\includegraphics[width=0.21\columnwidth]{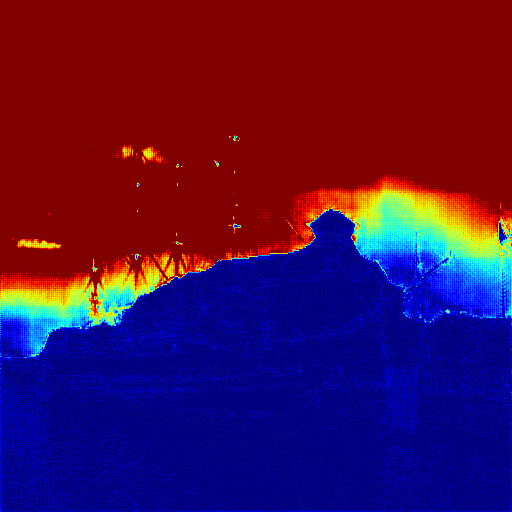}}
\subfigure[UNET+]{\centering\includegraphics[width=0.21\columnwidth]{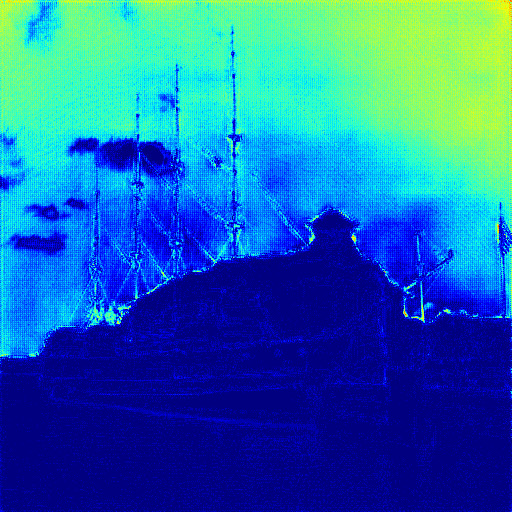}}
\subfigure[Target]{\centering\includegraphics[width=0.21\columnwidth]{figures/placeholder}}

\caption{Comparison on synthesized datasets. The two top samples are selected from S-COCO while the two bottom images come from S-Adobe5K. The methods with $+$ represent we replace the original coarser skip-connection with our S$^{2}$ASC. To visualize the differences better, we plot the absolute differences colormap between the harmonized image and target under each result. It is apparent that our method shows better results in both image quality and difference colormap. We enlarge the colormap 10$\times$ for better visualization.}
\vspace{-1.5em}
\label{fig:COCO}
\end{figure*}

\textbf{Deep Image Harmonization (DIH) \cite{Tsai:2017kv}} DIH harmonizes images in a supervised manner with an encoder-decoder structure. Additionally, they use the ground truth segmentation mask as a redundant decoder since the harmonized images should also share the same segmentation results with the target.
We train the DIH on the same synthesized datasets for fair comparison. As our S$^{2}$AM can be plugged into DIH easily by replacing the original skip-connections in color and semantic branch, we modify the original model for comparison on S-COCO and S-Adobe5K. Notice that, because S-Adobe5K has no ground truth segmentation mask, we only evaluate the color branch.

\textbf{RealismCNN (R-CNN)~\cite{Zhu:2015tl}} RealismCNN uses the pre-trained VGG16 network as the basic feature extractor and fine-tune the model in the fully-connection layer for classifying the realism of the input image. Additionally, they freeze the parameters in the realism model and optimize the color of the input image by a novel loss function. However, this method tries to solve a non-convex function by selecting the minimal cost from multiple running. We compare the results on our datasets with the official implementation\footnote{\hyperlink{https://github.com/junyanz/RealismCNN}{https://github.com/junyanz/RealismCNN}}.
%

\textbf{pix2pix~\cite{Isola:2016tp}} Pix2pix use GAN in the image-to-image transformation task firstly. GAN has significant benefits synthesizing the meaningful photo-realistic images because the discriminator in GAN distinguishes the real or fake image by the neural network itself. However, it fails to get the accurate numerical results. Since we use unet as our backbone network, we compare the unet, unet with S$^{2}$AM as Skip-connection~(Unet+S$^{2}$ASC) and unet with S$^{2}$AM as Decoder~(Unet+S$^{2}$AD). Furthermore, we compare pix2pix by considering Unet and Unet+S$^{2}$ASC as the generators of GAN respectively. Notice that, we train the GAN framework under the default configuration by the author's  implementation\footnote{\hyperlink{https://github.com/junyanz/pytorch-CycleGAN-and-pix2pix}{https://github.com/junyanz/pytorch-CycleGAN-and-pix2pix}}.

\subsection{Comparison}
\label{sec:com}

\subsubsection{Comparison on Synthesized Datasets.}

\begin{table}[t!]
\begin{center}
\begin{tabular}{|l|c|c|c|c|}
\hline
Method & MSE$\downarrow$ & SSIM$\uparrow$ & PSNR$\uparrow$  & LPIPS$\downarrow$ \\
\hline
\hline
\multicolumn{5}{|c|}{S-COCO dataset}  \\
\hline
Original Copy-and-Paste  & 25.94 & 0.9469 & 26.62 & 0.0507 \\
R-CNN \cite{Zhu:2015tl} & 31.21 &  0.9509 & 27.17 & 0.0458 \\
DIH~\cite{Tsai:2017kv}  &  20.88      &   0.9662     & 31.73 &  0.0308         \\
\textbf{DIH + S$^{2}$ASC}  &   18.09      &  0.9734      & 33.55    &   0.0174   \\
pix2pix~\cite{Isola:2016tp}    &   21.24   & 0.9633 & 32.13     & - \\
\textbf{pix2pix~\cite{Isola:2016tp} + S$^{2}$ASC}     &   17.67   & 0.9707 & 33.32  &   -  \\
Unet~\cite{Isola:2016tp}       & 17.28         & 0.9757          & 33.55 &        0.0162 \\
\textbf{Unet + S$^{2}$ASC } & 15.59 & \underline{0.9790} & 34.74 & \underline{0.0144} \\
\textbf{Unet + S$^{2}$AD} & \underline{14.88} & \underline{0.9790} & \underline{35.23} & \textbf{0.0128} \\
\textbf{Unet + S$^{2}$AD + pp} & \textbf{14.26} & \textbf{0.9811} & \textbf{35.38} & \textbf{0.0128} \\
\hline
\hline
\multicolumn{5}{|c|}{S-Adobe5K dataset}  \\
\hline
Original Copy-and-Paste  & 37.43  & 0.9401 & 26.60 & 0.0501 \\
R-CNN~\cite{Zhu:2015tl}  & 41.29 & 0.9338 & 26.11 & 0.0506 \\
DIH\cite{Tsai:2017kv}  & 34.71 &  0.9243 & 27.94 & 0.0506 \\
\textbf{DIH + S$^{2}$ASC}   & 26.96 & 0.9586 & 31.45  & 0.0282    \\
pix2pix~\cite{Isola:2016tp}      & 32.26       & 0.9560         & 30.01     &  - \\
\textbf{pix2pix~\cite{Isola:2016tp}  + S$^{2}$ASC}     &  22.67         & 0.9702         & 32.59    &   - \\
Unet~\cite{Isola:2016tp}   &     23.18 & 0.9693 & 32.46 & 0.0232   \\
\textbf{Unet + S$^{2}$ASC}  & 21.83 & 0.9711 & 33.31 & 0.0210 \\
\textbf{Unet + S$^{2}$AD} & \underline{21.52} & \underline{0.9743} & \underline{33.67} & \textbf{0.0186} \\
\textbf{Unet + S$^{2}$AD + pp} & \textbf{20.47} & \textbf{0.9757} & \textbf{33.94} & \underline{0.0195} \\
\hline
\end{tabular}
\end{center}

\caption{The numerical comparison on synthesized datasets. Here, S$^2$ASC and S$^2$AD are the different interpolate method of our S$^2$AM. \textbf{pp} represent the post-processing in Section.~\ref{sec:net}}.

\label{table:COCO}
\end{table}
For fair compare the effect of the backbone network and our attention module, we mainly compare the state-of-the-art methods on Unet+S$^{2}$ASC. Additionally, we compare the Unet+S$^{2}$ASC with Unet+S$^{2}$AD individually.

In S-COCO, we train all the learning-based methods under the same framework except the model structure. As shown in Table.~\ref{table:COCO}, RASC module get better numerical results when it is plugged into three baseline structures: DIH, unet and pix2pix. Moreover, our full method (Unet+S$^{2}$ASC) shows the best results. These experiments confirm that our method is more suitable for image harmonization task comparing with others because their approaches only gain the results relying on larger datasets. Additionally, our method achieves better visual effects. We plot the harmonized results in Figure.~\ref{fig:COCO} and calculate the absolute difference colormap between the target and result for better visualization. It is clear that our method outperforms other methods to a large extent. 
Figure.~\ref{fig:COCO}, our method shows better results in various image content and color form. It is obvious that our approach gains better results when the bed or person is the spliced region and performs well in both gray images and color images.

Similar improvements are also observed in S-Adobe5K. This dataset is more complicated than S-COCO for the data is limited, and the mask is inaccurate.
We show the visual comparisons in Figure.~\ref{fig:COCO} and report the numerical results in Table.~\ref{table:COCO}. Both results indicate that our methods gain better performances than others. As shown in the third example in Figure.~\ref{fig:COCO}, the three baseline methods with S$^{2}$AM perform better with inaccurate ground truth mask because our S$^{2}$AM filter the regional features separately in the skip-connection. 

\begin{figure}[t]
\centering
  \includegraphics[width=0.95\columnwidth]{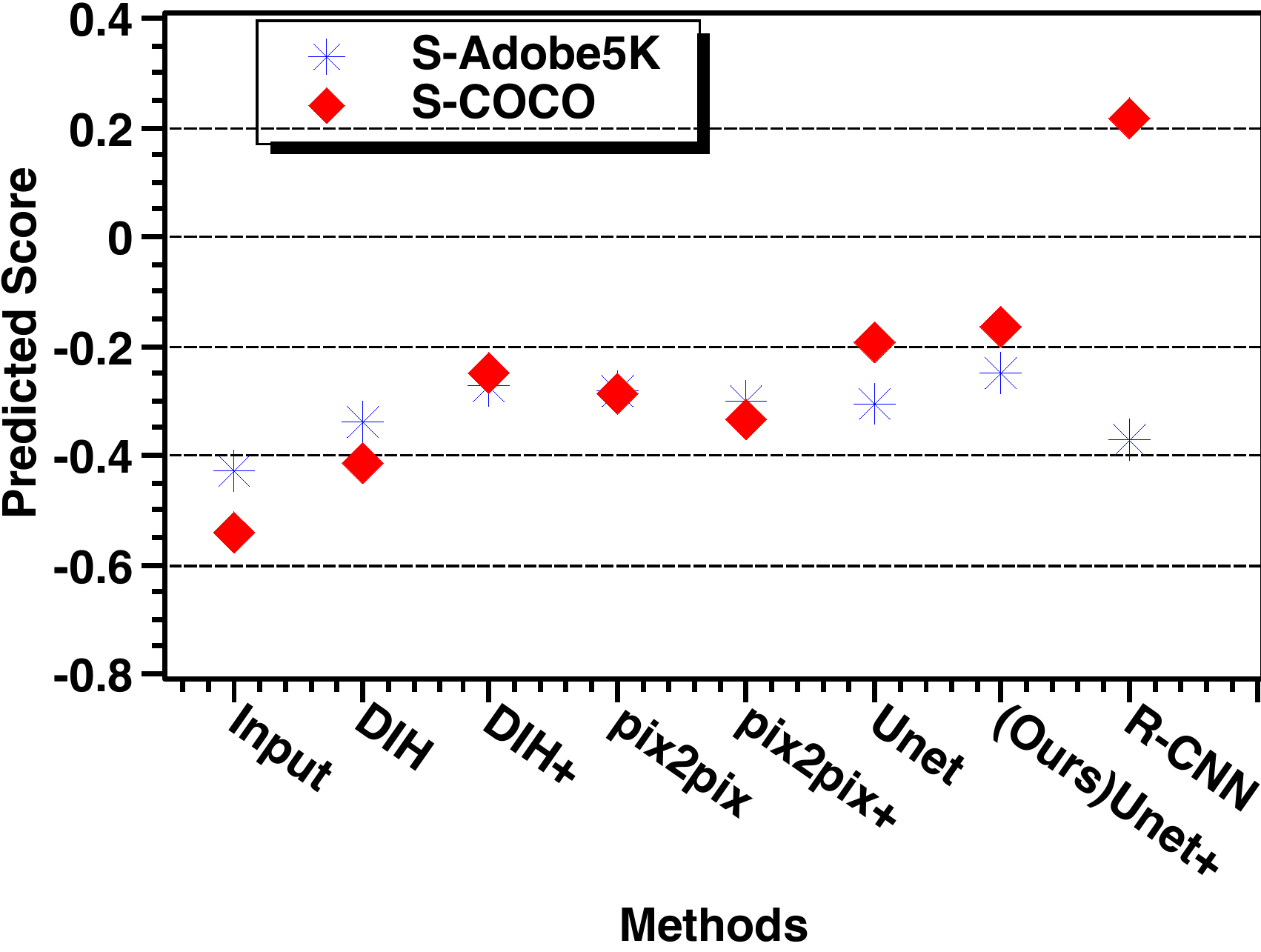}
  \caption{The predicted score on the pre-trained model of R-CNN~\cite{Zhu:2015tl}. The methods with $+$ represent we replace the original coarser skip-connection with our S$^2$ASC. A higher score means a more realistic result.}
  \label{figure:pretrained}
  \vspace{-1.5em}
\end{figure}

Besides the numerical and visual comparison, we also evaluate all the methods on the pre-trained composite realism model in R-CNN~\cite{Zhu:2015tl}. Notice that, the pre-trained model has not been trained on any images in our dataset before. The predicted score in Figure.~\ref{figure:pretrained} shows the predicted score of the composite images. It is clear that our S$^{2}$AM improves the predicted scores from the viewpoint of the pre-trained model in most cases. Meanwhile, our Unet+S$^{2}$AM method shows the best performance results comparing with others. However, it is not surprising that R-CNN performed better when the experiment is performed on its own pre-trained model. Interestingly, for R-CNN, we do not observe a similar phenomenon in S-Adobe5K. It is probably because the pre-trained model optimizes the difference in color by semantic while the mask is not always meaningful in S-Adobe5K.

\begin{figure}[t!]
\centering     
\subfigure{\centering\includegraphics[width=0.24\columnwidth,height=0.24\columnwidth]{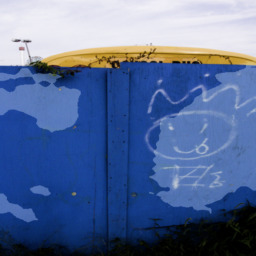}}
\subfigure{\centering\includegraphics[width=0.24\columnwidth]{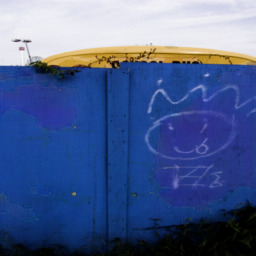}}
\subfigure{\centering\includegraphics[width=0.24\columnwidth]{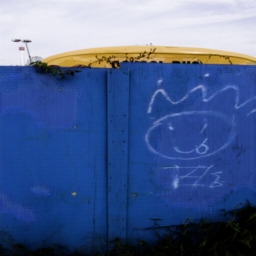}}
\subfigure{\centering\includegraphics[width=0.24\columnwidth]{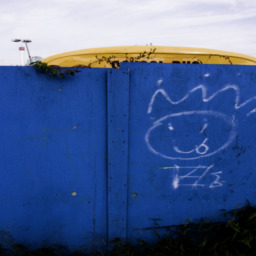}}
\par\bigskip\vspace*{-2em}

\subfigure{\centering\includegraphics[width=0.24\columnwidth,height=0.24\columnwidth]{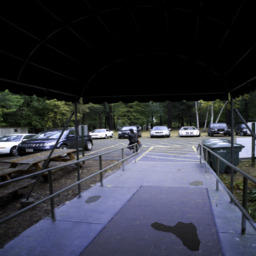}}
\subfigure{\centering\includegraphics[width=0.24\columnwidth]{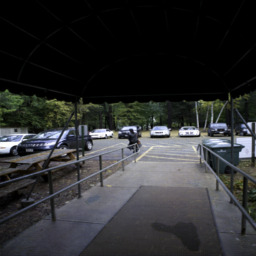}}
\subfigure{\centering\includegraphics[width=0.24\columnwidth]{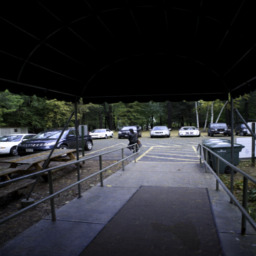}}
\subfigure{\centering\includegraphics[width=0.24\columnwidth]{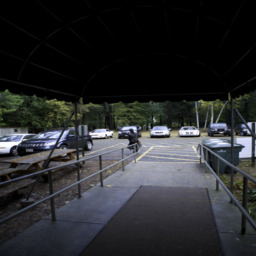}}
\par\bigskip\vspace*{-2em}

\subfigure{\centering\includegraphics[width=0.24\columnwidth,height=0.24\columnwidth]{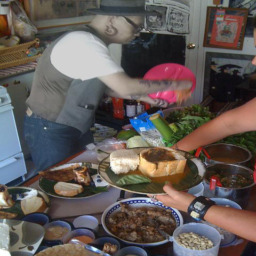}}
\subfigure{\centering\includegraphics[width=0.24\columnwidth,height=0.24\columnwidth]{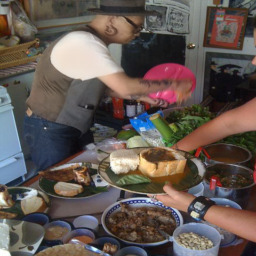}}
\subfigure{\centering\includegraphics[width=0.24\columnwidth]{figures/figsad/rascv2_000000182611_1}}
\subfigure{\centering\includegraphics[width=0.24\columnwidth]{figures/figsad/target_000000182611_1}}
\par\bigskip\vspace*{-2em}

\addtocounter{subfigure}{-12}
\subfigure[Input]{\centering\includegraphics[width=0.24\columnwidth,height=0.24\columnwidth]{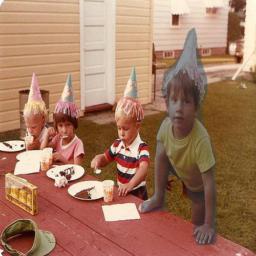}}
\subfigure[S$^2$ASC]{\centering\includegraphics[width=0.24\columnwidth]{figures/figsad/rascv1_000000345361_1.jpg}}
\subfigure[S$^2$AD]{\centering\includegraphics[width=0.24\columnwidth]{figures/figsad/rascv2_000000345361_1.jpg}}
\subfigure[Target]{\centering\includegraphics[width=0.24\columnwidth]{figures/figsad/target_000000345361_1.jpg}}

\caption{Comparison between the S$^2$ASC and S$^2$AD. Although S$^2$ASC harmonizes the tampered region, there are still some fake details while S$^2$AD show realistic global results.}
\label{fig:sad}
\vspace{-1.5em}
\end{figure}

We compare the results of S$^2$ASC and S$^2$AD on two synthesized datasets. We argue that, in Unet$+$S$^2$AD, our S$^{2}$AM learns from the high-level information and low-level information altogether. Besides the numerical analysis in Table.~\ref{table:COCO}, S$^2$AD show better global consistence that S$^2$ASC as shown in Fig.~\ref{fig:sad}. It is clear that S$^2$AD gets the benefits from the upsampled high-level features and generates more realistic results from the global context. For example, the shoulder of the human in the third example and the hat of the child in the last sample are not synthesized well in S$^2$ASC while we get the better results from the interpolate of high-level features and low-level information from encoder.

\subsubsection{Comparison on Real Dataset with User Study.}

\begin{figure}[t!]
\centering     

\subfigure{\centering\includegraphics[width=0.24\columnwidth,height=0.24\columnwidth]{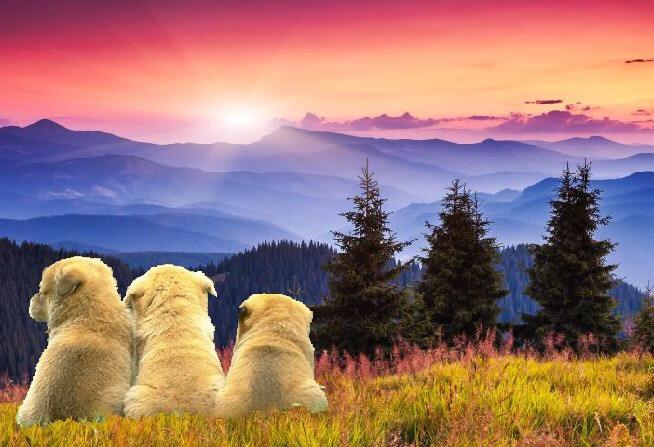}}
\subfigure{\centering\includegraphics[width=0.24\columnwidth]{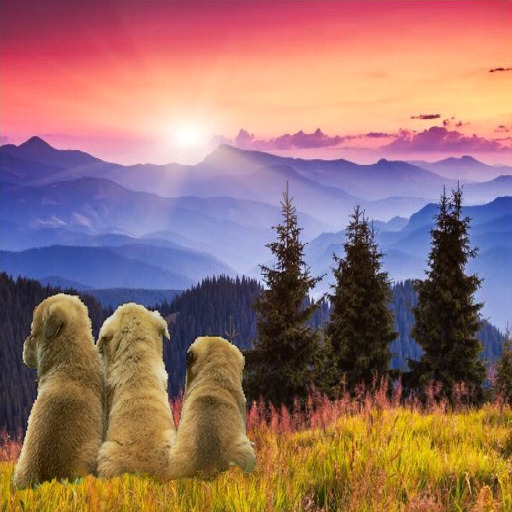}}
\subfigure{\centering\includegraphics[width=0.24\columnwidth]{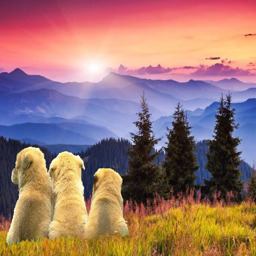}}
\subfigure{\centering\includegraphics[width=0.24\columnwidth]{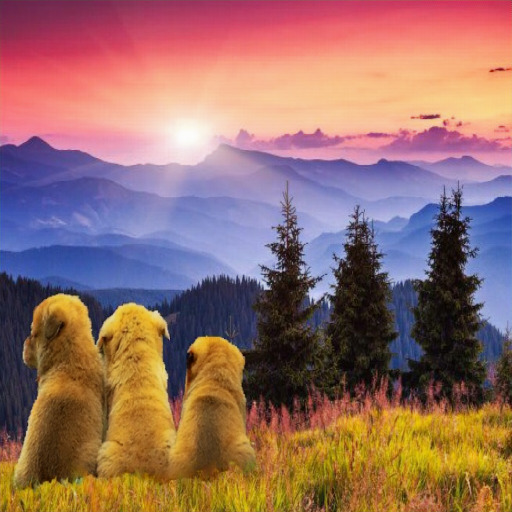}}
\par\bigskip\vspace*{-2em}

\subfigure{\centering\includegraphics[width=0.24\columnwidth,height=0.24\columnwidth]{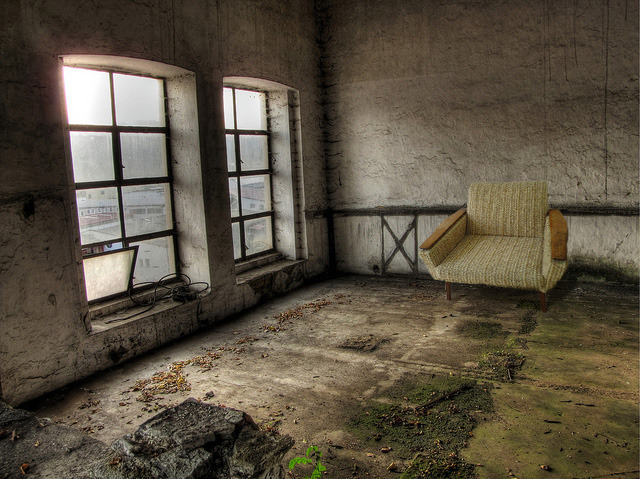}}
\subfigure{\centering\includegraphics[width=0.24\columnwidth,height=0.24\columnwidth]{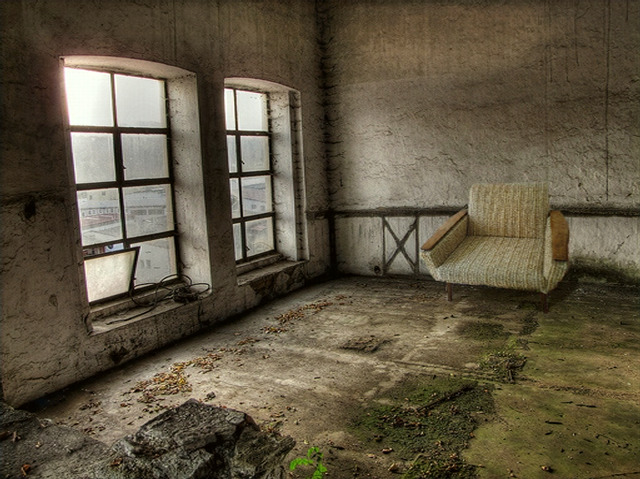}}
\subfigure{\centering\includegraphics[width=0.24\columnwidth]{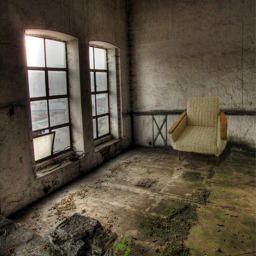}}
\subfigure{\centering\includegraphics[width=0.24\columnwidth]{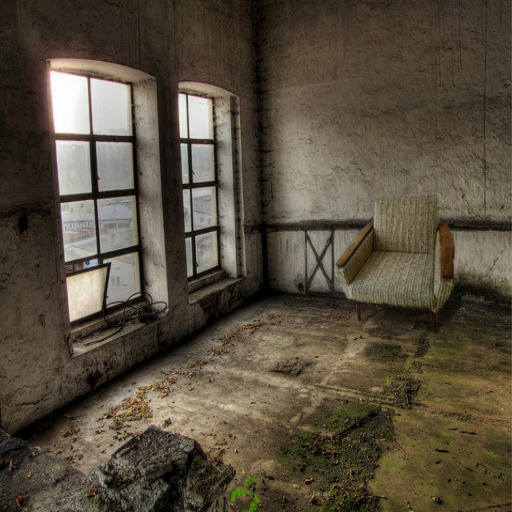}}
\par\bigskip\vspace*{-2em}

%

\addtocounter{subfigure}{-9}
\subfigure[Original]{\centering\includegraphics[width=0.24\columnwidth,height=0.24\columnwidth]{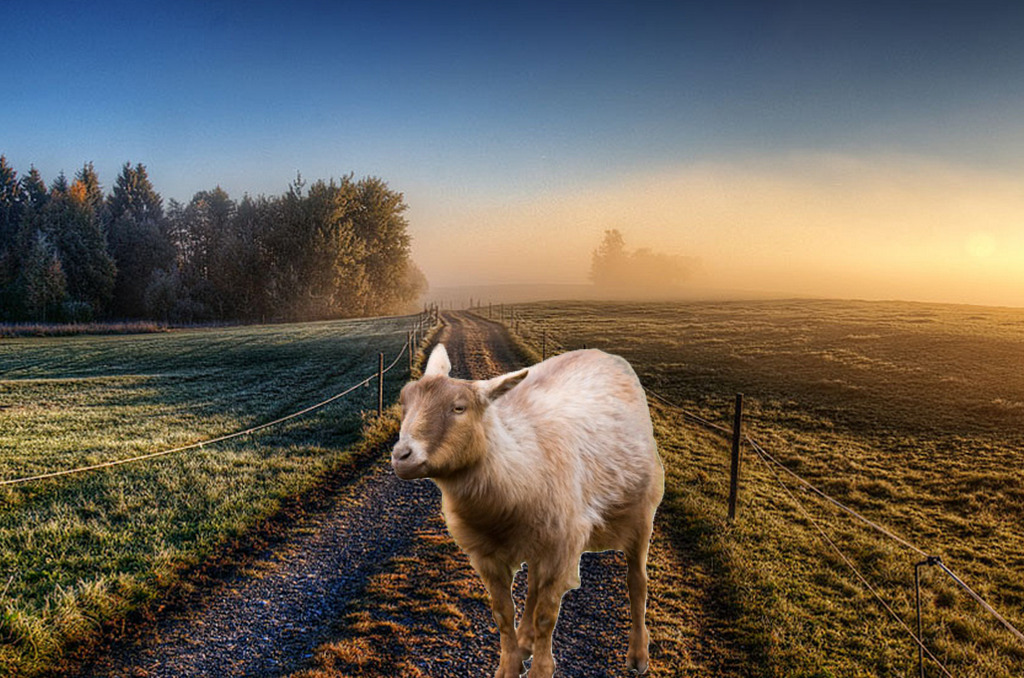}}
\subfigure[DIH~\cite{Tsai:2017kv}]{\centering\includegraphics[width=0.24\columnwidth,height=0.24\columnwidth]{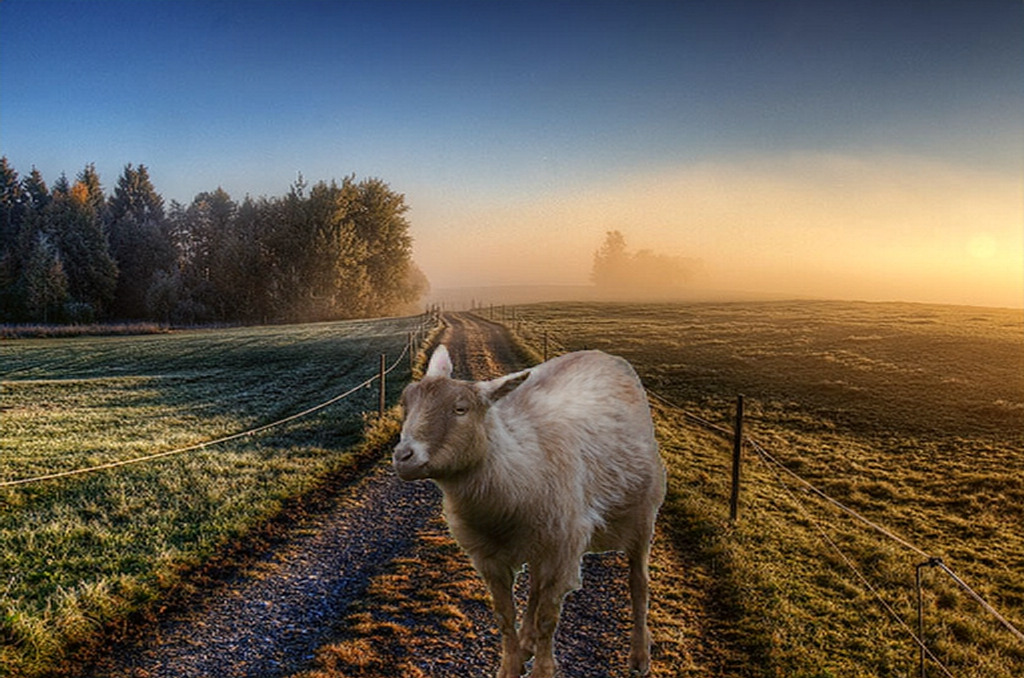}}
\subfigure[Zhu~\cite{Zhu:2015tl}]{\centering\includegraphics[width=0.24\columnwidth]{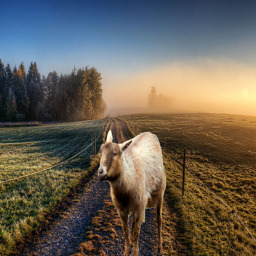}}
\subfigure[Ours]{\centering\includegraphics[width=0.24\columnwidth]{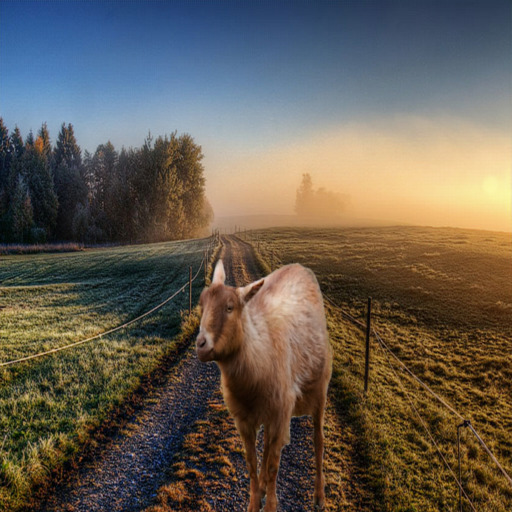}}
\caption{Comparison on the real dataset. As shown in the figure, our method gain better results than DIH~\cite{Tsai:2017kv} and Zhu~\cite{Zhu:2015tl} with the pre-trained model provided by the author.}
\label{fig:real}
\end{figure}
Although we train the model on the synthesized dataset, our method can also harmonize the real samples of the spliced image. We compare our method with the two most relevant state-of-the-art methods on the real composite images with provided spliced mask. This dataset contains 99 images with various different scenes and it is collected by DIH~\cite{Tsai:2017kv}. As shown in Figure.~\ref{fig:real}, our method out-perfumes other states-of-the-art methods with a larger margin on this dataset. For there is no available ground truth target image for the harmonized version of these images, we conduct a user study to evaluate our method on the real dataset and synthesized datasets by Amazon Mechanical Turk. Our user study template derivates from the web page in \cite{Wu:2017wq}. Specifically, for each task, giving the original copy-and-paste image and corresponding mask, the users need to choose the best realistic image from all the harmonized results created by different algorithms. In user study, we use the structure of Unet+S$^2$AD structure and train our model on two synthesized datasets with 30 epochs from scratch. As for the comparison, we compare our method with the pre-trained DIH model provided by author\footnote{\hyperlink{https://github.com/wasidennis/DeepHarmonization}{https://github.com/wasidennis/DeepHarmonization}}. This model pre-trained on the semantic segmentation task and fine-tune the network on three datasets synthesized by their own. As shown in the Table.~\ref{tab:ustudy}, our method gains much more votes than others with a larger margin.

\begin{table}[h]
\begin{center}
\begin{tabular}{|l|c|c|c|c|}
\hline
Method & Total votes   & Average votes  \\
\hline\hline
Zhu~\cite{Zhu:2015tl}  &  120  &   21\%   \\
DIH~\cite{Tsai:2016id} &  212  &   35\%  \\
Ours  &  263  &  44\%  \\
\hline

\end{tabular}
\end{center}

\caption{The comparison between our method and other state-of-the-art methods under user study.}

\label{tab:ustudy}

\end{table}

\subsection{Image Harmonization without mask}
\label{sec:comw}

\subsubsection{Comparison on Synthesized Dataset}

\begin{figure*}[h]
\centering     

\subfigure{\centering\includegraphics[width=0.28\columnwidth]{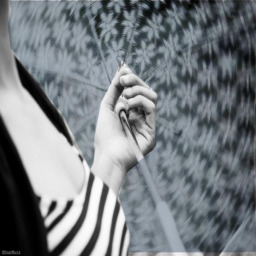}}
\subfigure{\centering\includegraphics[width=0.28\columnwidth]{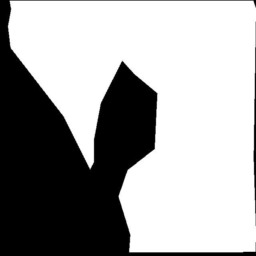}}
\subfigure{\centering\includegraphics[width=0.28\columnwidth]{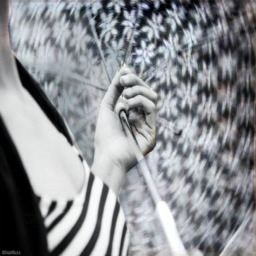}}
\subfigure{\centering\includegraphics[width=0.28\columnwidth]{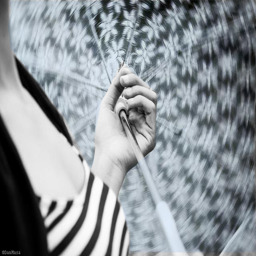}}
\subfigure{\centering\includegraphics[width=0.28\columnwidth]{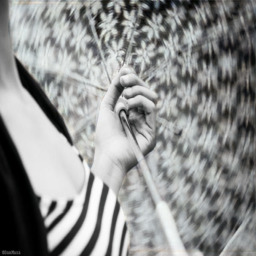}}
\subfigure{\centering\includegraphics[width=0.28\columnwidth]{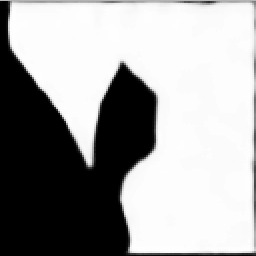}}
\subfigure{\centering\includegraphics[width=0.28\columnwidth]{figures/coco/target_180}}

\bigskip
\vspace{-2em}

\subfigure{\centering\includegraphics[width=0.28\columnwidth]{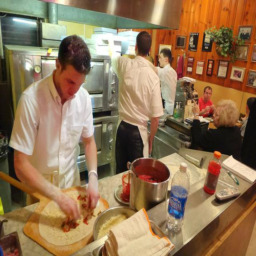}}
\subfigure{\centering\includegraphics[width=0.28\columnwidth]{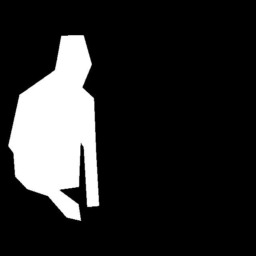}}
\subfigure{\centering\includegraphics[width=0.28\columnwidth]{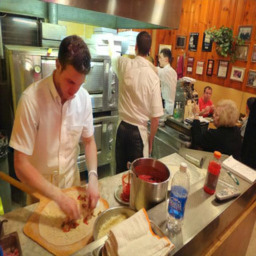}}
\subfigure{\centering\includegraphics[width=0.28\columnwidth]{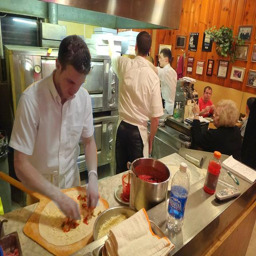}}
\subfigure{\centering\includegraphics[width=0.28\columnwidth]{figures/coco/_mmaskedgannaivemmucrossx_952}}
\subfigure{\centering\includegraphics[width=0.28\columnwidth]{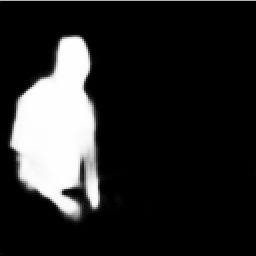}}
\subfigure{\centering\includegraphics[width=0.28\columnwidth]{figures/coco/target_952}}

\bigskip
\vspace{-2em}

\subfigure{\centering\includegraphics[width=0.28\columnwidth]{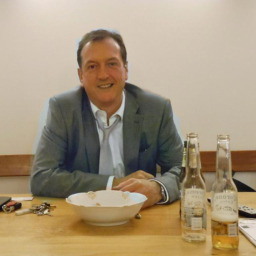}}
\subfigure{\centering\includegraphics[width=0.28\columnwidth]{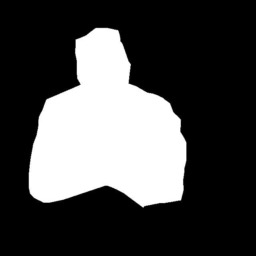}}
\subfigure{\centering\includegraphics[width=0.28\columnwidth]{figures/coco/_basicnaiveuno_149}}
\subfigure{\centering\includegraphics[width=0.28\columnwidth]{figures/coco/zhu_149}}
\subfigure{\centering\includegraphics[width=0.28\columnwidth]{figures/coco/_mmaskedgannaivemmucrossx_149}}
\subfigure{\centering\includegraphics[width=0.28\columnwidth]{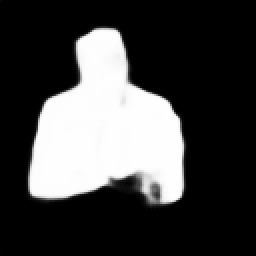}}
\subfigure{\centering\includegraphics[width=0.28\columnwidth]{figures/coco/target_149}}

\bigskip
\vspace{-2em}
\addtocounter{subfigure}{-21}

\subfigure[Input]{\centering\includegraphics[width=0.28\columnwidth]{figures/coco/input_905}}
\subfigure[Tampered Mask]{\centering\includegraphics[width=0.28\columnwidth]{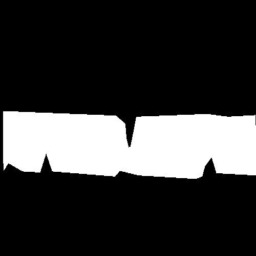}}
\subfigure[Unet~\cite{Isola:2016tp}]{\centering\includegraphics[width=0.28\columnwidth]{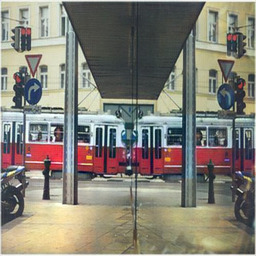}}
\subfigure[Zhu~\cite{Zhu:2015tl}]{\centering\includegraphics[width=0.28\columnwidth]{figures/coco/zhu_905}}
\subfigure[Ours]{\centering\includegraphics[width=0.28\columnwidth]{figures/coco/_mmaskedgannaivemmucrossx_905}}
\subfigure[Attention Map]{\centering\includegraphics[width=0.28\columnwidth]{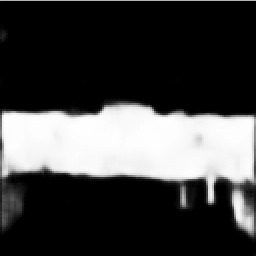}}
\subfigure[Target]{\centering\includegraphics[width=0.28\columnwidth]{figures/coco/target_905}}

\caption{The Comparison results of our methods and other baseline methods on the S-COCO datasets without the ground truth mask as input. We also plot the attention map created by our method and the ground truth map here for comparison.}

\label{fig:unmask}
\vspace{-1.5em}
\end{figure*}

As described in Section.~\ref{sec:net}, our method can be adapted to the image harmonization task without the ground truth as input. So in this task, we regard the previous Unet as the baseline network structure by feeding the color image to the network only. We evaluate our method on the S-COCO dataset for comparison. As shown in Fig.~\ref{fig:unmask}, even without the feeding of the tampered mask, our method can still predict the realistic harmonized image and achieve much more better performance than the baseline unet~\cite{Isola:2016tp} both visual and numerical quality. We also plot the intermediate attention map (coarser level of the attention loss) created by our methods in Fig.~\ref{fig:unmask}. The mask predicted by our method is reasonable. We also report the numerical results under this task in Table.~\ref{tab:umask}, surprisedly, our full method get comparable results with the DIH methods (which need the mask as input) under the same datasets. Table.~\ref{tab:umask} also report the intermediate results of our methods. Each experiment shows the necessity in the network.

\begin{table}[h]
\begin{center}
\begin{tabular}{|l|c|c|c|c|}
\hline
Method & MSE$\downarrow$ & SSIM$\uparrow$ & PSNR$\uparrow$ \\

\hline\hline
Original Copy-and-Paste  & 25.94 & 0.9469 & 26.62 \\
Unet~(w/o mask) ~\cite{Isola:2016tp}  &  29.86     &    0.9518     &   28.96  \\
Ours w/o Attention Loss &  26.08 & 0.9608 & 30.10  \\
Ours w/o GAN Loss        &   23.93   &  0.9622     &   30.89  \\
Ours w/o post-processing    & 23.86 & 0.9590 & 30.61 \\
Ours (Unet+S$^{2}$AD)  &   \textbf{20.73}    &    \textbf{0.9657}     &  \textbf{31.15}  \\
\hline
\end{tabular}
\end{center}

\caption{We evaluate the proposed method for image harmonization without the ground truth mask as input on the S-COCO datasets.}
\label{tab:umask}
\vspace{-1.5em}
\end{table}

\subsubsection{Evaluation of Attention Loss}
\begin{figure}[h]
  \includegraphics[width=\columnwidth]{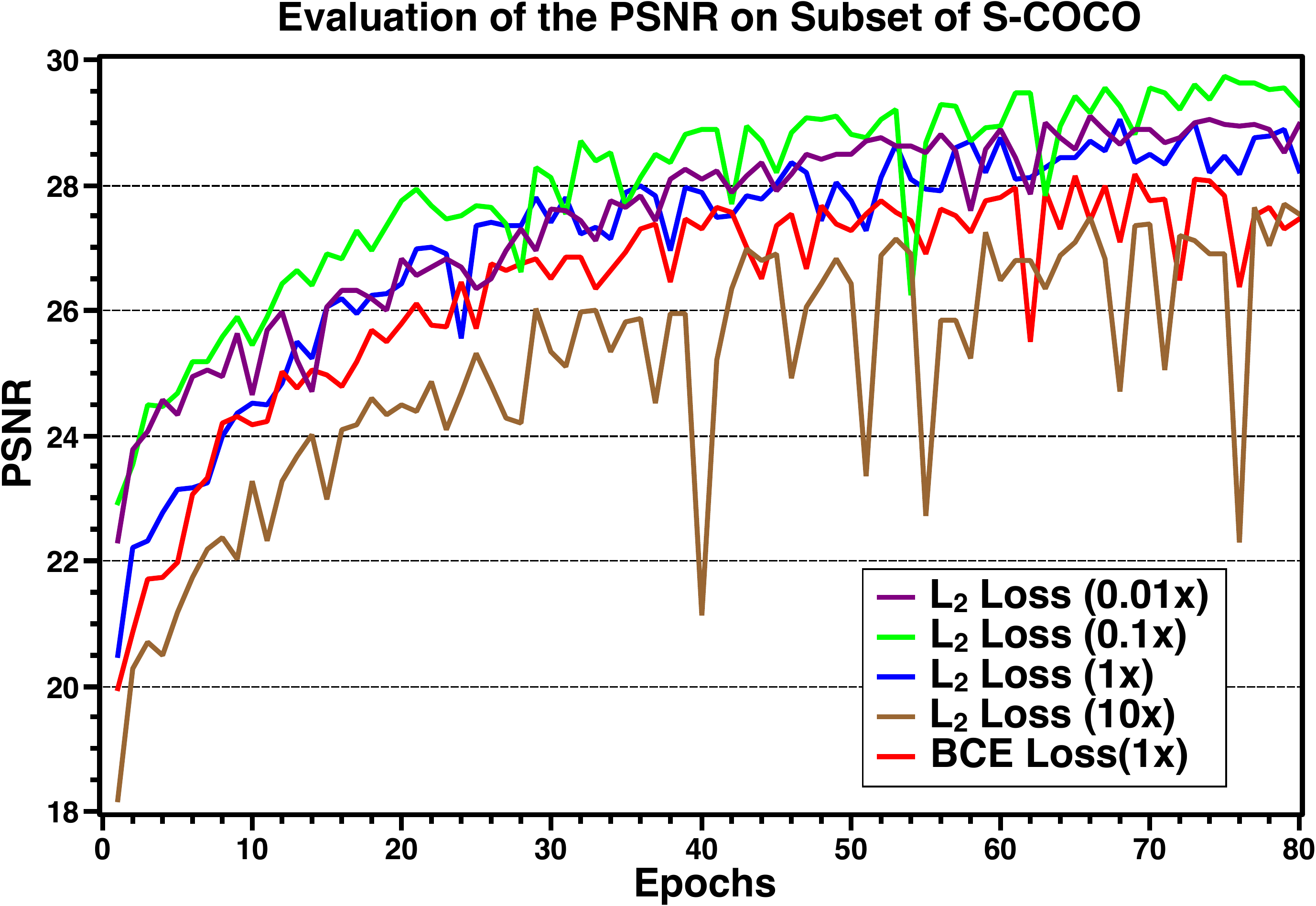}
  \caption{The Evaluations of the attention loss on the Subset of S-COCO. We plot the PSNR on the full test set in each epoch when we train the network with different attention loss. We also tune the hyper-parameters $\alpha,\beta$ in Equation.\ref{eq:att}. $X\times$ means the proportion between the pixel level loss parameters $\alpha$ and our attention loss $\beta$. This figure is best view in color.}
    \label{fig:attpsnr}
    \vspace{-1.5em}
\end{figure}

The key component of our method for image harmonization without the mask is attention loss. We tuning the choice of attention loss of $L_2$, \textit{Binary Cross Entropy} under a subset of S-COCO dataset. This subset contains 10k training images and we test it on the same test dataset as the full S-COCO dataset. We choose this subset S-COCO for evaluating the attention loss because it trains 4$\times$ faster than the original dataset. As shown in the red line and blue line in Fig.~\ref{fig:attpsnr}, $L_2$ loss is a more suitable choice for attention loss than \textit{Binary Cross Entropy} under the same ratio between the pixel loss. We also try to find the best proportion by rescaling the percentage of the attention loss to 0.1$\times$, 0.01$\times$,10$\times$. From the Fig.~\ref{fig:attpsnr}, it is clear that the best proportion between attention loss and pixel-level loss is $1:10$.

\subsection{Attention Mechanisms Analysis}
\label{sec:attention}
We first evaluate the proposed S$^2$AM against other state-of-the-art attention mechanisms: \textit{Squeeze-and-Excitation Network~(SE-Block)}~\cite{Hu:2017tf} and \textit{Convolution Block Attention Module~(CBAM)}~\cite{Woo:2018wr} for image harmonization task in Section.~\ref{sec:evaatt}. Their methods are proposed for image classification tasks, so we replace the original skip-connection in Unet for comparison with the proposed S$^{2}$ASC. Besides, we increase the channels in each levels of the original Unet to match the parameters of our Unet+S$^2$AM model for a fair comparison~($+$ More channels in Table.~\ref{table:evaluation}). Finally, we compare the attention modules with the basic CONV-Block. Each CONV-Block block has the structure of CONV~($3\times3$)-ELU-BN-CONV~($3\times3$)-ELU-BN, which is the same as the learning-able block in S$^{2}$AM. Then, we conduct a detailed ablation study of our module in Section.~\ref{sec:astudy} and Section.~\ref{sec:attaly}. All the methods are experimented on S-COCO dataset in the same framework with default configurations.


\subsubsection{Evaluation of Attention Modules}
\label{sec:evaatt}

\begin{figure}[h]
\centering   
\subfigure[Input]{\centering\includegraphics[width=0.30\columnwidth]{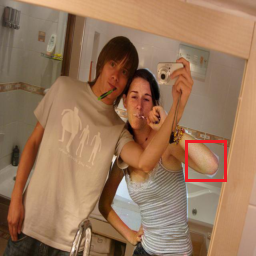}}
\subfigure[Mask]{\centering\includegraphics[width=0.30\columnwidth]{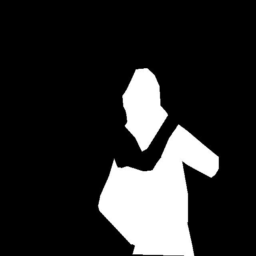}}
\subfigure[Target]{\centering\includegraphics[width=0.30\columnwidth]{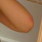}}

\par\bigskip\vspace*{-2em}
\subfigure{\centering\includegraphics[width=0.30\columnwidth]{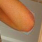}}
\subfigure{\centering\includegraphics[width=0.30\columnwidth]{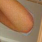}\label{fig:attention_cbam}}
\subfigure{\centering\includegraphics[width=0.30\columnwidth]{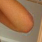}\label{fig:attention_rasc}}

\par\bigskip\vspace*{-2em}
\addtocounter{subfigure}{-3}
\subfigure[SE-Block\cite{Hu:2017tf}]{\centering\includegraphics[width=0.30\columnwidth]{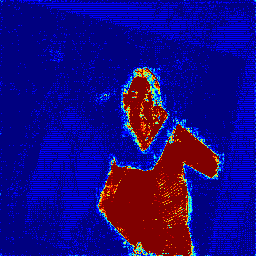}}
\subfigure[CBAM\cite{Woo:2018wr}]{\centering\includegraphics[width=0.30\columnwidth]{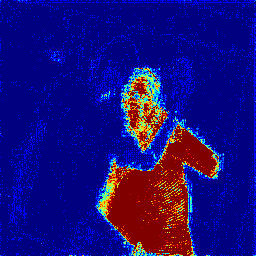}}
\subfigure[S$^{2}$AM~(Ours)]{\centering\includegraphics[width=0.30\columnwidth]{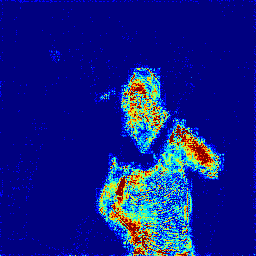}}

\caption{Comparisons on different attention modules. We plot the results of Unet+SE-Block, Unet+CBAM and Unet+S$^{2}$AM~(Ours) for comparison. For each method, we show the harmonized results of the red region and the absolute difference colormap between the result and target image. It is clear that our Unet + S$^{2}$AM method gains better results than others in the boundary of arms (best view with zoom in) and the absolute difference colormaps. Moreover, our method also shows a better prediction in the non-spliced region. The colormaps are enlarged 30$\times$ for better visualize the differences in the non-spliced region.}
\label{fig:attentionmodule}
\vspace{-1.5em}
\end{figure}

As illustrated in Table.~\ref{table:evaluation}, Unet+CONV-Block and Unet+SE-Block show poor performance than the baseline method for similar reasons. In these methods, the convolution operators are performed on the whole image and their method cannot learn the disparities in the spliced region and preserve the features in the non-spliced region simultaneously. SE-Block shows better results than CONV-Block because SE-Block learns the channel attention additionally in the skip-connection.
Moreover, it is not surprising that CBAM gets better results than the baseline since the spatial attention part in CBAM can learn the specific region automatically, and the input mask of our network makes it easy.
However, our unet+S$^{2}$ASC method gets a better result than other attention modules with a large margin. This is because, with the help of the hard-coded mask, we can design more channel attention modules for different purposes and learning specifically while the SE-Block and CBAM can only focus on certain channels or certain regions by learning from data. Besides, the baseline network with more parameters also improve the performance, however, it still far worse than our method. Overall, the benefits of our attention module are obvious. From Fig.\ref{fig:attentionmodule}, our method gets a cleaner difference colormap in the background than other attention modules for their methods only consider  soft attention. Another improvement between Figure.\ref{fig:attention_cbam} to Figure.\ref{fig:attention_rasc} is the boundary of the spliced mask. From the figure, we can find our method gets better harmonized edges while the ground mask is not always true while other methods fail. It is probably because the soft spatial attentions in CBAM can not always get an accurate attention map even with the input ground truth mask while we use the hard-coded mask and Gaussian Filter in the module. 



\begin{table}[h]
\begin{center}
\begin{tabular}{|l|c|c|c|c|}
\hline
Method & MSE$\downarrow$ & SSIM$\uparrow$ & PSNR$\uparrow$ \\
\hline\hline
baseline unet~\cite{Isola:2016tp}  & 17.28         & 0.9757          & 33.55        \\
~~$+$ More Channels  & 16.93  & 0.9770  & 33.80  \\
~~$+$ CONV-Block   & 17.85 &  0.9757 & 33.34  \\
~~$+$ SE-Block~\cite{Hu:2017tf} & 17.45 & 0.9762 & 33.42 \\
~~$+$ CBAM~\cite{Woo:2018wr}  & 16.50 & 0.9768 & 34.32 \\
\textbf{unet with S$^{2}$ASC (Ours)} & 15.59 & \textbf{0.9790} & \textbf{34.74} \\
 ~~$ - $ $G_{bg}, G_{mix}, G_{fg}$ & 17.93 & 0.9758 & 33.32 \\
 ~~$ - $ $G_{mix}$ & 15.80 & 0.9779 & 34.49 \\
 ~~$ - $ Gaussian Mask & 15.67 & 0.9788 & 34.72 \\
 ~~$+$ 6 layers  S$^{2}$ASC  & 15.66 & 0.9787 & 34.73 \\
 ~~$ - $ Learning block($L$) & \textbf{15.49} & 0.9781 & 34.73  \\
\hline
\end{tabular}
\end{center}

\caption{The evaluation of S$^{2}$AM with the different attention modules and the ablation study of S$^{2}$AM on S-COCO dataset.}
\label{table:evaluation}
\vspace{-2em}
\end{table}

\subsubsection{Ablation Study}
\label{sec:astudy}
We conduct ablation experiments on the inner structure of the proposed S$^{2}$AM model under \textit{\textbf{S$^{2}$ASC}}. All the experiments are performed on the S-COCO dataset with the same configuration. We illustrate all the numerical results in Table.~\ref{table:evaluation}.

\textbf{Without $G_{bg}, G_{mix}, G_{fg}$.} {By comparison with the baseline, Unet w/o $G_{bg}, G_{mix}, G_{fg}$ gets slightly worse results because the hard-coded mask will force the learning-able block $L$ to learn the necessary and unnecessary features altogether without any identification.}


\textbf{Without $G_{mix}$.} We remove the $G_{mix}$ in S$^{2}$ASC module for necessity. As shown in Table.~\ref{table:evaluation}, compared with S$^{2}$ASC, the method without $G_{mix}$ performs slightly worse because $G_{mix}$ selects the original features in the spliced region which is unnecessary to transfer.

\begin{figure}[h!]
\centering     
\subfigure[w/o Gaussian Filter]{\centering\includegraphics[width=0.45\columnwidth]{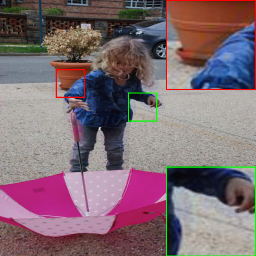}}
\subfigure[w/  Gaussian Filter]{\centering\includegraphics[width=0.45\columnwidth]{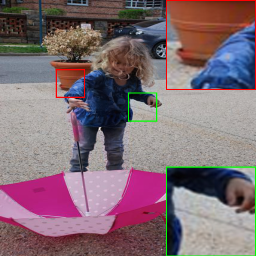}}
\caption{Gaussian Filter smooth the binary mask to fit object. For example, the little girl in the image is the tampered region with inaccuracy spliced mask. From two zoom-in regions in the image, the method with Gaussian Filter show better harmonization boundary.}
\label{fig:gaussian}
\vspace{-1.5em}
\end{figure}

\textbf{Without Gaussian Filter.} A detailed explanation of Gaussian Filter has been introduced in the Section.\ref{sec:rasc}. Here, we report the results of experiments. As shown in Fig.~\ref{fig:gaussian}, the Gaussian Filter shows better performance in the boundary of spliced edges. We also report the numerical results in Table.~\ref{table:evaluation}, it is obvious that the Gaussian Filter improves the performance of our module.

\textbf{With 6 layers S$^{2}$ASC.} The model gets a slightly worse performance when we  replace all the skip-connection with S$^{2}$ASC in Unet (Unet $+$ 6 layers S$^{2}$ASC in Table.~\ref{table:evaluation}). This experiment also fit our assumption: the style is much more related to low-level features, and the high-level features should be the same. However, in all 6 layers S$^{2}$ASC, our method still gains much better results than the baseline. 


\textbf{Without learning-able Block.} We evaluate the importance of our learning-able block. As shown in Table.\ref{table:evaluation}, the methods without learning-able block $L$ show worse results than our full method, especially in SSIM. This is because SSIM measures the image quality from the luminance, contrast, and structure while the small convolutions in the learning-able block will learn more detail. 

\subsubsection{Attention Analysis}
\label{sec:attaly}

\begin{figure}[h]
	\centering
	\includegraphics[width=\columnwidth]{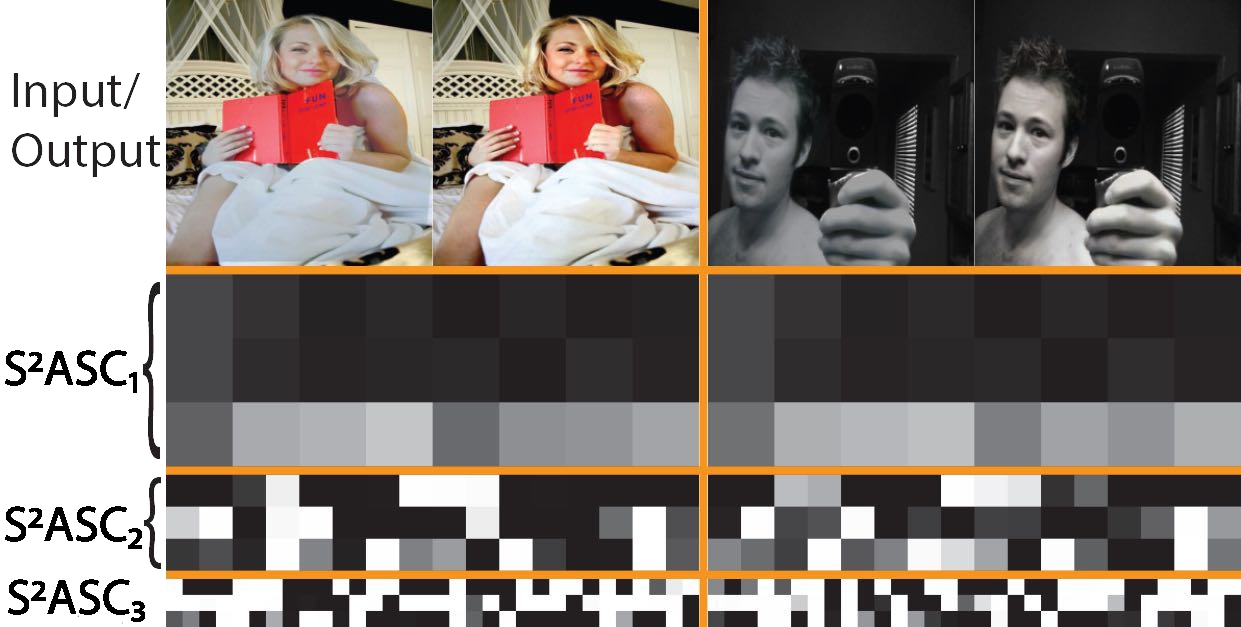}

	\caption{From coarse-to-fine, we plot the first $\frac{1}{8}$ attention channels response on three different levels of S$^{2}$ASC in network structure named $S^2ASC_1$, $S^2ASC_2$ and $S^2ASC_3$, respectively. For each level of attention, we visualize the Channel Attention Module with the order $G_{bg}$, $G_{mix}$ and $G_{fg}$ from the top to the bottom.}
	\label{figure:attention_analysis}

\end{figure}

We also analyze the detail responses of all three \textit{Channel Attention Modules} in our S$^{2}$AM modules. We use \textbf{\textit{S$^{2}$ASC}} in unet structure for evaluating for the features used in S$^{2}$ASC all comes from the encoder. As shown in Fig.~\ref{figure:attention_analysis}, 
we visualize the weighting of $G_{fg}, G_{mix}, G_{bg}$ in the three layers of S$^{2}$ASC modules at the bottom of each sample. 
Particularly, the blacker block indicates the channel weighting is low while the whiter block means it is important for filtered features. So our first observation is that our S$^{2}$ASC module allocates and weights the features according to the input images by comparing the response maps between two images. Moreover, Another interesting observation is that there are more white blocks in $G_{fg}$ than $G_{mix}$ and $G_{bg}$ in the $S^2ASC_1$ module while the opposite phenomenon is shown in $S^2ASC_3$. This fact perfectly explains our assumption: the coarser features in the spliced region need to learn specifically while the high-level features are almost the same.

\section{Conclusion}

In this paper, we propose a novel method for image harmonization. Specifically, we design a new attention module called S$^{2}$AM to fit the image harmonization task with Unet backbone network. Additionally, we harmonize the image without mask by interpolating spatial attention module and the attention loss to S$^{2}$AM module. This idea comes from the original intention of the similarity in content and difference in the spliced region between the spliced image and harmonized target. The experiments show that our proposed method achieves better results than others both in quality and quantity. 

Besides image harmonization, the proposed attention module can also be adapted to other computer vision tasks easily with regional differences, such as \textit{Free-Form Image Inpainting}~\cite{Yu:2018uw} and \textit{Semantic Image Synthesis}~\cite{Park:2019uq}. We believe it is a promising direction for our future work.


\section*{Acknowledgments}
The authors would like to thank Nan Chen and Xuebo Liu for helpful discussion. This work was partly supported by the University of Macau under Grants: MYRG2018-00035-FST and MYRG2019-00086-FST, and the Science and Technology Development Fund, Macau SAR (File no. 041/2017/A1, 0019/2019/A).

\ifCLASSOPTIONcaptionsoff
  \newpage
\fi

{
\bibliographystyle{IEEEtran}
\bibliography{IEEEabrv}
}

\end{document}